\documentclass{article} 
\usepackage{iclr2024_conference,times}

\usepackage{hyperref}
\usepackage{url}

\usepackage{booktabs}       
\usepackage{amsfonts}       
\usepackage{nicefrac}       
\usepackage{microtype}      
\usepackage{xcolor}         

\usepackage{adjustbox}
\usepackage{diagbox}
\usepackage{multirow}
\usepackage{makecell}
\usepackage{subfigure}

\usepackage{times}
\usepackage{epsfig}
\usepackage{graphicx}
\usepackage{amsmath}
\usepackage{amssymb}
\usepackage{colortbl}

\usepackage[switch]{lineno}
\usepackage{bbding}
\usepackage{ragged2e}
\usepackage{float}
\usepackage{booktabs}

\title{ViDA: Homeostatic Visual Domain Adapter for Continual Test Time Adaptation}

\author{Jiaming Liu\textsuperscript{\rm 1}, 
Senqiao Yang\textsuperscript{\rm 1\thanks{Equal contribution, \textsuperscript{\Envelope} Corresponding author.}},
Peidong Jia \textsuperscript{\rm 1}, 
Renrui Zhang \textsuperscript{\rm 3}, \\
Ming Lu\textsuperscript{\rm 1},
Yandong Guo\textsuperscript{\rm 2},
Wei Xue\textsuperscript{\rm 4},
Shanghang Zhang\textsuperscript{\rm 1}~\textsuperscript{\Envelope}
\vspace{0.1cm}\\
\textsuperscript{\rm 1}National Key Laboratory for Multimedia Information Processing,\\ School of Computer Science, Peking University
\textsuperscript{\rm 2}AI$^{2}$Robotics \\ \textsuperscript{\rm 3} The Chinese University of Hong Kong \textsuperscript{\rm 4} Hong Kong University of Science and Technology\\
jiamingliu@stu.pku.edu.cn, yangsenqiao.ai@gmail.com, shanghang@pku.edu.cn
}

\iclrfinalcopy 
\begin{document}

\maketitle

\begin{abstract}
Since real-world machine systems are running in non-stationary environments, Continual Test-Time Adaptation (CTTA) task is proposed to adapt the pre-trained model to continually changing target domains. Recently, existing methods mainly focus on model-based adaptation, which aims to leverage a self-training manner to extract the target domain knowledge. However, pseudo labels can be noisy and the updated model parameters are unreliable under dynamic data distributions, leading to error accumulation and catastrophic forgetting in the continual adaptation process. To tackle these challenges and maintain the model plasticity, we design a Visual Domain Adapter (ViDA) for CTTA, explicitly handling both domain-specific and domain-shared knowledge. Specifically, we first comprehensively explore the different domain representations of the adapters with trainable high-rank or low-rank embedding spaces. Then we inject ViDAs into the pre-trained model, which leverages high-rank and low-rank features to adapt the current domain distribution and maintain the continual domain-shared knowledge, respectively. To exploit the low-rank and high-rank ViDAs more effectively, we further propose a Homeostatic Knowledge Allotment (HKA) strategy, which adaptively combines different knowledge from each ViDA. Extensive experiments conducted on four widely used benchmarks demonstrate that our proposed method achieves state-of-the-art performance in both classification and segmentation CTTA tasks. Note that, our method can be regarded as a novel transfer paradigm for large-scale models, delivering promising results in adaptation to continually changing distributions. Project page: \href{https://sites.google.com/view/iclr2024-vida/home}{https://sites.google.com/view/iclr2024-vida/home}
\end{abstract}

\section{Introduction}
Deep Neural Networks (DNN) have achieved remarkable performance in various computer vision tasks, such as classification~\citep{he2016deep}, object detection~\citep{zhu2020deformable}, and segmentation~\citep{xie2021segformer} when the test data distribution is similar to the training data. However, real-world machine perception systems~\citep{yang2023lidar, li2023manipllm, arnold2019survey} operate in non-stationary and constantly changing environments, which contain heterogeneous and dynamic domain distribution shifts. Applying a pre-trained model in these real-world tasks \citep{sakaridis2021acdc} can lead to significant degradation in perception ability on target domains, especially when the target distribution changes unexpectedly over time. 
Therefore, the development of continual domain adaptation (DA) methods is essential for enhancing the generalization capability of DNNs and improving the reliability of machine perception systems in dynamic environments.

A classical source-free DA task, Test-Time Adaptation \citep{liang2023comprehensive} (TTA), eases the distribution shift between a source domain and a fixed target domain. This is typically achieved through the utilization of self-training mechanisms \citep{mummadi2021test, wang2020tent}. 
However, when adapting to continually changing target domains, pseudo labels are noisy and the updated model parameters become uncertain, leading to error accumulation and catastrophic forgetting. 
To tackle this problem, Continual Test-Time Adaptation (CTTA) has been proposed \citep{wang2022continual}, which addresses a sequence of different distribution shifts over time rather than a single shift as in TTA.
Furthermore, CTTA also encompasses the efficient continual adaptation of foundation models \citep{kirillov2023segment} to continual downstream tasks or distributions \citep{Bahngetal2022}.

\begin{figure*}[t]
    \vspace{-0.3cm}
    \centering
    \includegraphics[width=0.99\linewidth]{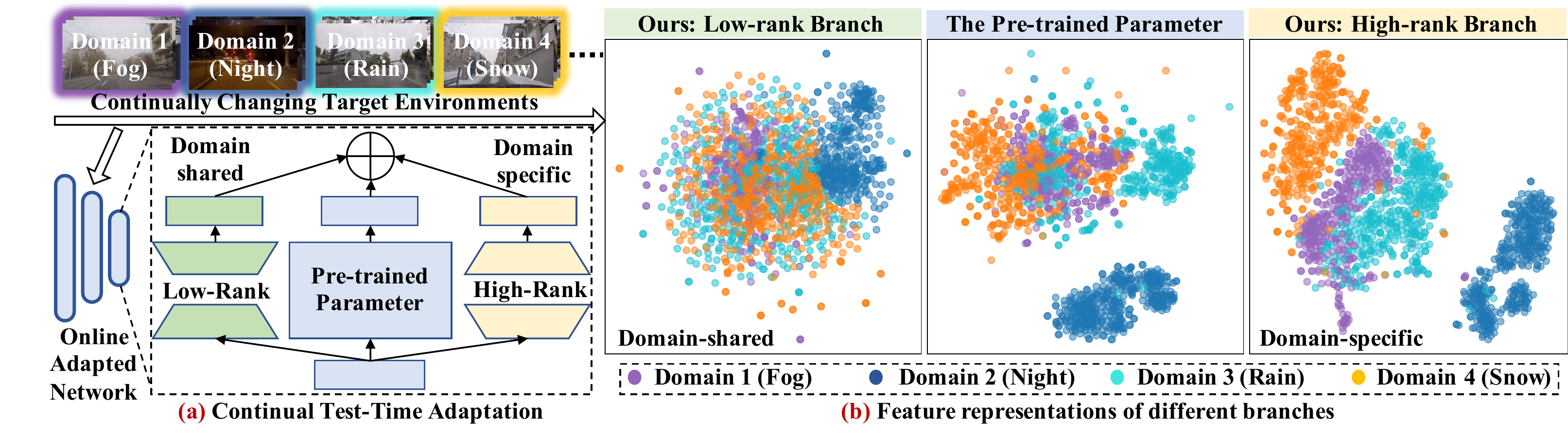}
    \vspace{-0.3cm}
    \caption{
    \textbf{The problem and motivation.}
    \textcolor{red}{(a)} Our goal is to effectively adapt the source pre-trained model to continually changing target domains. We propose Visual Domain Adapters with high-rank and low-rank embedding spaces to tackle the error accumulation and catastrophic forgetting challenges during the continual adaptation process.
    \textcolor{red}{(b)} we conduct a t-SNE \citep{van2008visualizing} distribution analysis for the different adapter representations across four target domains (ACDC). The low-rank branch exhibits a consistent distribution across the target domains, suggesting that it can effectively disregard the impact of dynamic distribution shifts. The high-rank branch demonstrates noticeable distribution discrepancies between the various target domains, suggesting that it primarily focuses on extracting domain-specific knowledge. 
    }
    \label{intro}
    \vspace{-0.39cm}
\end{figure*}

Existing CTTA works have primarily employed model-based or prompt-based approaches to extract target domain-specific and domain-shared knowledge simultaneously. However, for model-based methods \citep{wang2022continual, chakrabarty2023sata}, the noisy pseudo labels are still unreliable and play a limited role in avoiding error accumulation, particularly in scenarios with significant distribution gaps. Meanwhile, prompt-based methods \citep{gan2023decorate, yang2023exploring} face difficulties in leveraging soft prompts with limited trainable parameters to learn long-term domain-shared knowledge and prevent catastrophic forgetting.

To tackle these limitations and maintain the model plasticity, we design a homeostatic Visual Domain Adapter (ViDA), shown in Fig .\ref{intro} (a), which explicitly manages domain-specific and domain-shared knowledge in the continual adaptation process.
Specifically, we first carefully explore the different domain representations of ViDAs with trainable high or low-dimension embedding space in the middle layer. 
As shown in Fig. \ref{intro} (b), our observations reveal that ViDA with a low-rank embedding space focuses on task-relevant feature representation, showing trivial distribution distance in different domains and neglecting the influence of dynamic distribution shifts.
Conversely, ViDA with a high-rank feature concentrates more on extracting domain-specific knowledge, as evidenced by the feature distribution in different target domains showing an obvious discrepancy. 
We provide a detailed explanation of the motivations in Section~\ref{sec:motivation} and Appendix~\ref{ap:mo}.

This observation motivates us to inject ViDAs into the pre-trained model, which leverages different domain representations of high and low-dimension features to avoid error accumulation and catastrophic forgetting simultaneously.
To better extract different domain knowledge, we further propose a Homeostatic Knowledge Allotment (HKA) strategy to dynamically fuse the knowledge from low-rank and high-rank ViDA. Based on the data distribution, HKA adaptively regularizes the balance of different feature representations, including original model, domain-specific, and task-relevant features.
During inference, the low-rank and high-rank ViDAs can be projected into the pre-trained model by re-parameterization \citep{ding2021repvgg}, which ensures no extra parameter increase and maintains the model plasticity. 
In summary, our contributions are as follows:

\textbf{1)} We study the different domain representations of the adapters with high-rank and low-rank features. Then we design a Visual Domain Adapter (ViDA), explicitly managing domain-specific and task-relevant knowledge to tackle the error accumulation and catastrophic forgetting problem, respectively.

\textbf{2)}
Considering the various distribution shifts for each target sample, we further propose a Homeostatic Knowledge Allotment (HKA) strategy to dynamically merge knowledge from low-rank and high-rank ViDAs, thus enhancing ViDAs' distinct domain representations.

\textbf{3)}
Our proposed approach outperforms most state-of-the-art methods according to the experiments on four benchmark datasets, covering classification and segmentation CTTA. 

\textbf{4)}
Our CTTA method provides a novel transfer paradigm for large-scale models, delivering promising results in adaptation to continually changing distributions. Meanwhile, we empower the source model with domain generalization ability through the proposed homeostatic ViDAs, achieving a significant improvement on the unseen target domains.

\section{Related work}

\textbf{Test-time adaptation (TTA)}, also referred to as source-free domain adaptation~\citep{Boudiafetal2023, Kunduetal2022, ShiqiYangetal2021}, aims to adapt a source model to an unknown target domain distribution without relying on any source domain data. Recent research has explored self-training and entropy regularization techniques to fine-tune the source model ~\citep{Liangetal2020, Chenetal2022}. Tent~\citep{DequanWangetal2021} updates the training parameters in batch normalization layers by minimizing entropy. 
This approach has prompted subsequent exploration in recent works~\citep{niu2023towards, yuan2023robust}, which continue to investigate the robustness of normalization layers.

\textbf{Continual Test-Time Adaptation (CTTA)} refers to a scenario where the target domain is not static, presenting additional challenges for traditional TTA methods. The first approach to address this challenging task is introduced in ~\citep{wang2022continual}, which combines bi-average pseudo labels and stochastic weight reset. For addressing error accumulation, Ecotta \citep{song2023ecotta} introduces a meta-network to regularize the outputs from both the meta-network and the frozen network. 
And RMT \citep{dobler2023robust} introduces a symmetric cross-entropy loss.
While these works tackle the CTTA problem at the model level, ~\citep{gan2023decorate,yang2023exploring, ni2023distribution} utilize visual domain prompts or a small fraction of parameters to extract continual target domain knowledge.

\textbf{Parameter-Efficient Fine-Tuning (PEFT)} has gained significant traction within the field of natural language processing (NLP) ~\citep{ hu2021lora, houlsby2019parameter, zaken2021bitfit, hu2022knowledgeable, gao2021making, he2021effectiveness, vu2022spot, qin2021exploring}. Adapter-based models, a form of PEFT, have gained popularity in NLP. They employ bottleneck architecture adapter modules inserted between layers in pre-trained models.
Inspired by NLP, adapters in visual tasks have also received widespread attention. 
In the initial phases of adapter development, residual adapter modules~\citep{rebuffi2017learning, rebuffi2018efficient} are proposed to aid in the effective adaptation of convolutional neural networks across multiple downstream tasks. 
AdaptFormer ~\citep{chen2022adaptformer} enhances the ViT \citep{dosovitskiy2020image} model by replacing the original multi-layer perceptron (MLP) block with a down-to-up bottleneck module in a parallel manner.
VL-Adapter~\citep{sung2022vl} improves the efficiency and performance of adapters by sharing low-dimensional layer weights to attain knowledge across tasks.

\section{Method}

\textbf{Preliminary.} In Continual Test-Time Adaptation (CTTA), we pre-train the model $q_\theta(y|x)$ on the source domain $D_{S}={(Y_S, X_S)}$ and adapt it on multiple target domains $D_{T_i}={\{(X_{T_i})\}}_{i=1}^{n}$, where $n$ represents the scale of the continual target datasets. The entire process can not access any source domain data and can only access target domain data once. The distributions of the target domains (i.e., ${D_{T_1}, D_{T_2}, ..., D_{T_n}}$) are constantly changing over time. Our goal is to adapt the pre-trained model to target domains and maintain the perception ability of the model on the seen domain distributions.

\begin{figure*}[t]
\centering
\vspace{-0.3cm}
\includegraphics[width=\linewidth]{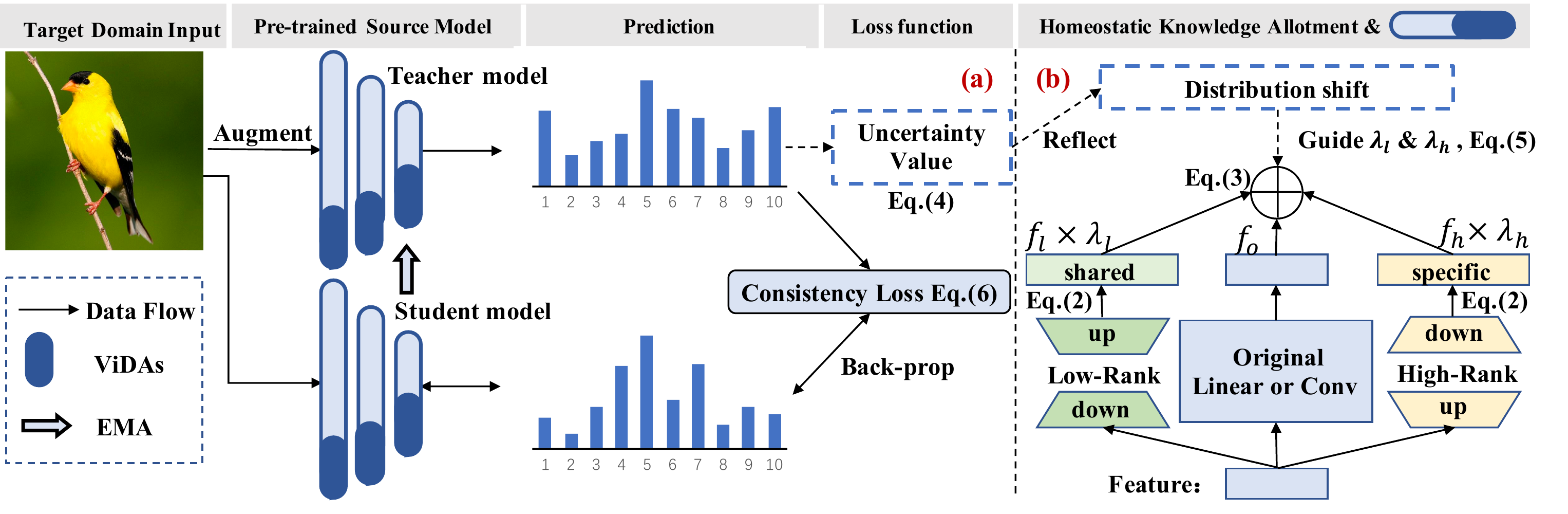}
\vspace{-0.7cm}
\caption{\textbf{The framework of Visual Domain Adapter (ViDA).} \textcolor{red}{(a)} We inject low-rank and high-rank ViDAs into either linear or Conv layers of the pre-trained source model. The student model processes the original image, while the teacher model processes an augmented version of the same image.
To update the ViDAs, we construct a teacher-student framework and use a consistency loss (Eq. \ref{eq:loss}) as the optimization objective. 
In addition, the teacher model calculates an uncertainty value (Eq. \ref{eq:mc}), reflecting the distribution shift of each sample in target domains. \textcolor{red}{(b)} Based on the degree of distribution shift, we introduce the Homeostatic Knowledge Allotment (HKA) strategy, which aims to dynamically fuse the knowledge from each ViDA with different domain representation.}
\label{fig:framework} 
\vspace{-0.2cm}
\end{figure*}

\textbf{Overall Framework.} 
Drawing from the insight that mean teacher predictions are often more robust than standard models \citep{tarvainen2017mean, dobler2023robust}, we utilize a teacher-student framework to ensure stability during continual domain adaptation, which also presents a fair comparison with previous CTTA works \citep{wang2022continual, gan2022decorate}. The overall framework and the details of our method are shown in Fig .\ref{fig:framework}.

\subsection{Motivation}
\label{sec:motivation}

The CTTA encounters significant challenges, primarily due to error accumulation and catastrophic forgetting \citep{wang2022continual}. Meanwhile, adapters with different dimensional middle-layer features demonstrate effectiveness in addressing these challenges. This encourages us to take a step further and justify the principles underlying the use of low-rank adapter and high-rank adapter in the CTTA.

\textbf{Low-rank adapter.}
Our hypothesis regarding the effectiveness of adapters in mitigating catastrophic forgetting is that their low-rank embedding space representation plays a crucial role. To explore this further, we conduct a t-SNE distribution study \citep{van2008visualizing} on the third transformer block to analyze the feature distributions across four target domains (ACDC dataset~\citep{sakaridis2021acdc}). The results are depicted in Fig. \ref{intro} (b). Our analysis reveals that the low-rank adapter exhibits a relatively consistent distribution across the different target domains, suggesting that its low-rank embedding space can effectively disregard the impact of dynamic distribution shifts and prioritize the extraction of domain-shared knowledge.

\begin{figure*}[t]
\centering
\vspace{-0.3cm}
\includegraphics[width=0.85\linewidth]{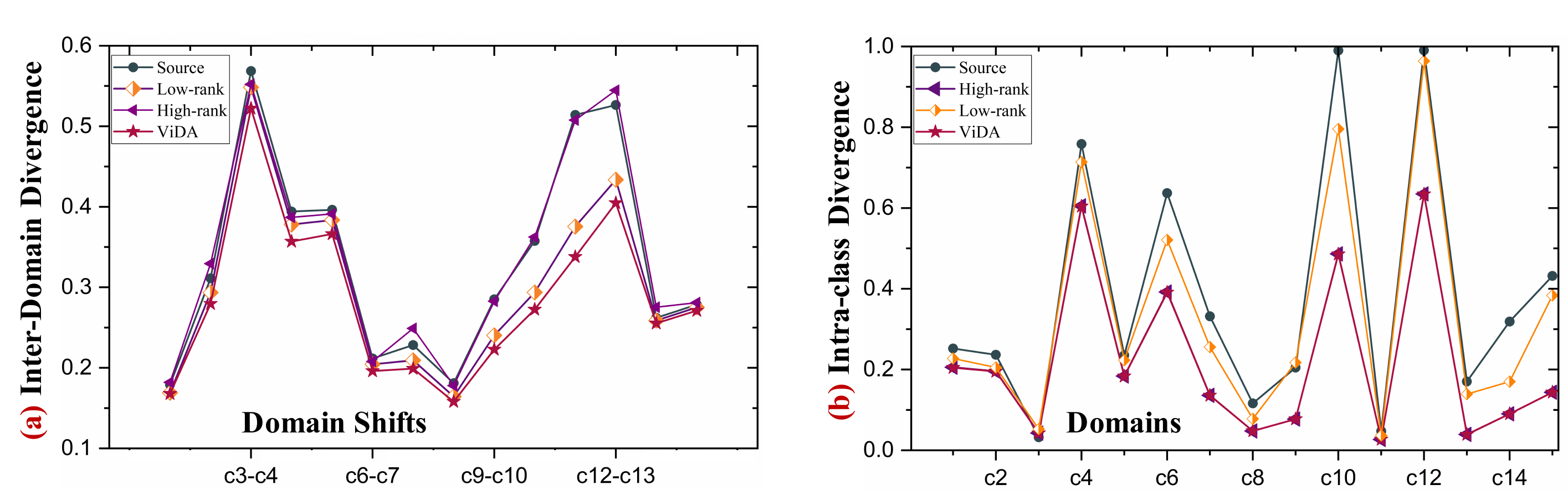}
\vspace{-0.3cm}
\caption{c1 to c15 represent the 15 corruption domains in CIFAR10C listed in sequential order. \textcolor{red}{(a)} Low-rank adapter based model effectively mitigates inter-domain divergence than the source model across all 14 domain shifts. \textcolor{red}{(b)} High-rank adapter based model significantly enhances the intra-class feature aggregation, yielding results that closely approximate those achieved by our ViDA method.}
\label{motivation}
\end{figure*}

Furthermore, we adopt the domain distance definition proposed by Ben-David \citep{ben2006analysis, ben2010theory} and build upon previous domain transfer research \citep{ganin2016domain} by employing the $\mathcal{H}$-$divergence$ metric to evaluate the domain representations of adapters across different target domains. The discrepancy distance between two distributions $D_S$ and $D_{T_i}$ can be calculated as: 
\begin{equation}
\vspace{-0.2cm}
d_\mathcal{H}(D_S, D_{T_i}) = 2 \mathop{\mathrm{sup}}_{\mathcal{D} \sim \mathcal{H}} | \mathop{\mathrm{Pr}}_{x \sim D_S}[\mathcal{D}(x)=1] - \mathop{\mathrm{Pr}}_{x \sim D_{T_i}}[\mathcal{D}(x)=1] |
\label{eq:dh}
\vspace{-0.05cm}
\end{equation}
, where $\mathcal{H}$ denotes hypothetical space and $\mathcal{D}$ denotes discriminator.
Similar to \citep{ruder2017learning,allaway2021adversarial}, we adopt the $Jensen$-$Shannon\ (JS)\ divergence$ between two adjacent domains as an approximation of $\mathcal{H}$-divergence because it has been shown to successfully distinguish domains.
If the inter-domain divergence is relatively small, it can be demonstrated that the feature representation is consistent and less influenced by cross-domain shifts \citep{ganin2016domain}. We compare the $JS$ values obtained by using the source model alone, injecting a low-rank adapter, injecting a high-rank adapter, and combining low- and high-rank adapters, as illustrated in Fig. \ref{motivation} (a). Our results indicate that the feature representation generated by the low-rank adapter exhibits lower divergence compared to both the original source model and the high-rank adapter, especially when dealing with later target domains or significant domain shifts between adjacent domains (i.e., target domains 9-13). This result simultaneously demonstrates the low-rank adapter's ability to learn long-term domain-shared knowledge in a continually changing environment.

\begin{figure*}[t]
\vspace{-0.2cm}
\centering
\includegraphics[width=0.99\linewidth]{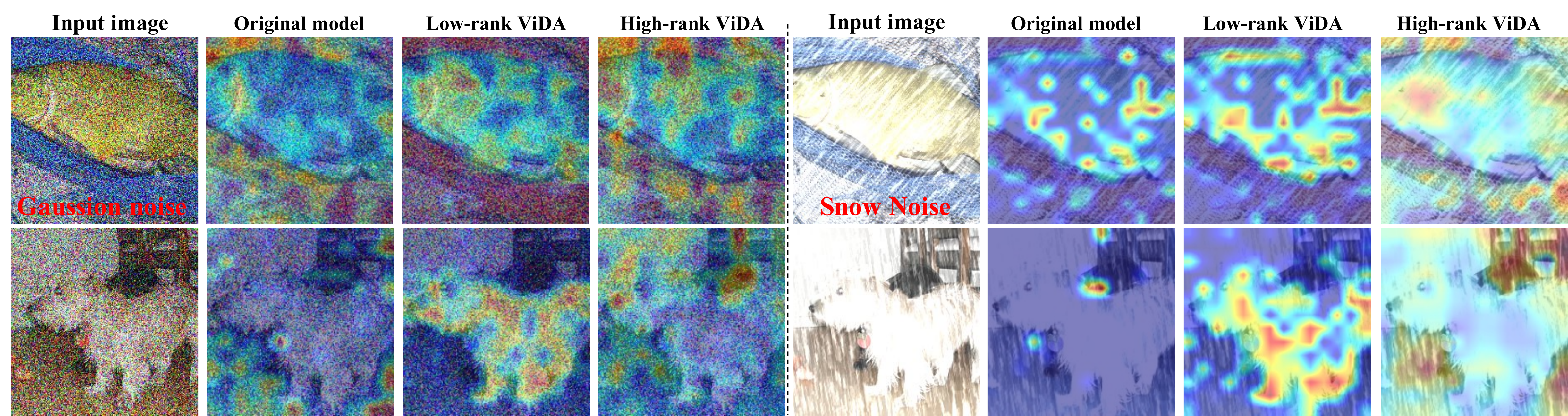}
\vspace{-0.2cm}
\caption{The qualitative analysis of the CAM. We adopt CAM to compare the attention of the low-rank branch, high-rank branch, and the original model during the continual adaptation process.}
\label{fig: cam}
\vspace{-0.3cm}
\end{figure*}

To provide clearer evidence for the intuition, we extend our analysis by incorporating the qualitative analysis of Class Activation Mapping (CAM) on the ImageNet-to-ImageNet-C CTTA. As shown in Fig .\ref{fig: cam}, we showcase the feature representations from different target domains, including the noise of Gaussian and Snow. We observe that the low-rank ViDA is inclined to put more weight on the foreground sample while tending to disregard background noise shifts. This indicates that the low-rank ViDA attends to locations with more general and task-relevant information.

\textbf{High-rank adapter.} 
Regarding the domain representation of the adapter with a high-rank feature, we propose that it is better suited to address error accumulation in the continual adaptation process. We verify this by t-SNE analyzing the feature distributions between different domains, as shown in Fig. \ref{intro} (b), and observe that there is a clear discrepancy between domains. The distribution achieves a better aggregation in a single domain. This suggests that high-ranking adapters have a better grasp of target domain data distribution.
Inspired by intra-class dissimilarity proposed by $k$-means~\citep{macqueen1967classification}, we use normalized intra-class divergence to further verify the domain representations of high-rank adapters in CIFAR10C.
In a given domain, if the intra-class divergence for each category is smaller, it demonstrates that the model has a better understanding of the current distribution \citep{li2020maximum}.
As illustrated in Fig. \ref{motivation} (b), the high-rank adapter is found to drive down the intra-class divergence within almost all domains, indicating that it can better adapt to current domain distribution and extract domain-specific knowledge in continual target domains.
For more straightforward verification, we conduct qualitative analysis by incorporating the visualization of CAM. Conversely, the high-rank ViDA exhibits an inverse pattern, as illustrated in Fig .\ref{fig: cam}. It allocates more attention to locations characterized by substantial domain shift, encompassing the entirety of the input images. This behavior aligns with the high-rank branch's tendency to fit global information and predominantly extract domain-specific knowledge from the target domain data.

In conclusion, the structure of low-rank ViDA reduces feature redundancy, which leads to an underfit state during CTTA. Consequently, it tends to acquire general information across continuous target domains, extracting task-relevant knowledge to mitigate catastrophic forgetting. In contrast, high-rank ViDA employs a higher-dimensional feature representation that better aligns with the target data distribution, thereby focusing on learning domain-specific knowledge to prevent error accumulation. We offer additional justifications and specifically designed experiments in Appendix~\ref{ap:mo}.

\subsection{Visual Domain Adapter}
The above observation motivates us to introduce high-rank and low-rank Visual Domain Adapters (ViDAs) into the source pre-trained model, aiming to simultaneously adapt current domain distribution and maintain the continual domain-shared knowledge in CTTA. 

\noindent\textbf{The architecture.} The design principle of injecting ViDAs into the pre-trained model is simple yet effective, which is illustrated in Fig .\ref{fig:framework} (b). As we can see there are three sub-branches, the linear (or Conv) layer in the middle branch is originated from the original network, while the right branch and left branch are bottleneck structures and separately indicate the high-rank ViDA and low-rank ViDA. Specifically, the right branch (high-rank) contains an up-projection layer with parameters $W^h_{up} \in R^{d \times d_{h}}$, a down-projection layer with parameters $W^h_{down} \in R^{d_{h} \times d}$, where $d_{h}$ (e.g., $d_{h} = 128$) is the middle dimension of high-rank feature and satisfies $d_{h} \geq d$. There is not any non-linear layer in the ViDA. And we utilize the linear layer as the projection layer when the original model is transformer architecture and adopt $1 \times 1$ Conv as the projection layer when the original model is a convolution network. In contrast, the left low-rank branch first injects a down-projection layer with parameters $W^l_{down} \in R^{d \times d^{l}}$, then place an up-projection layer with parameters $W^l_{up} \in R^{d_{l} \times d}$, where $d_{l}$ (e.g., $d_{l} = 1$) stand for the middle dimension of the low-rank feature ($d_{l} \ll d$).  For a input feature $f$, the produced features of high-rank ($f_h$) and low-rank ViDA ($f_l$) are formulated as: 
\begin{equation}
 f_{h} = W^h_{down} \cdot (W^h_{up} \cdot f);\quad   f_{l} = W^l_{up} \cdot (W^l_{down} \cdot f)
\label{eq:feature}
\end{equation}
The two-branch bottleneck is connected to the output feature of the original network ($f_o$) through the residual connection via scale factors ($\lambda_h$ and $\lambda_l$). The fusion knowledge ($f_f$) can be described as:
\begin{equation}
 f_{f} = f_{o} + \lambda_h \times f_{h} + \lambda_l \times f_{l} 
\label{eq:fusion}
\end{equation}
The domain knowledge scale factors ($\lambda_h$ and $\lambda_l$) are adaptively obtained through the homeostatic knowledge allotment strategy, which is shown in Section \ref{sec:HKA}.
During inference, the different domain-represented ViDAs (linear relation) can be projected into the pre-trained model by re-parameterization \citep{ding2021repvgg}, which ensures no extra model parameter increase of the original model.

\subsection{Homeostatic Knowledge Allotment}
\label{sec:HKA}

\noindent\textbf{Method motivation.} 
In CTTA, target domain data can only be accessed once and exhibits different distribution shifts, which underscores the importance of efficient domain transfer. Moreover, to effectively address error accumulation and catastrophic forgetting, it becomes necessary to extract different knowledge and manage it separately. Although the specialized structures of low-rank and high-rank ViDAs contribute to distinct domain representation learning, the continual adaptation process also needs to regularize the knowledge fusion weight to ensure the efficient capture of relevant domain-specific knowledge without compromising the retention of long-term domain-shared knowledge.
\noindent\textbf{HKA design.} 
As depicted in Fig .\ref{fig:framework} \textcolor{red}{(b)}, we draw inspiration from \citep{ovadia2019can, roy2022uncertainty} and introduce an uncertainty value to quantify the degree of distribution shift for each sample. While the confidence score is a common measure to assess prediction reliability, it tends to fluctuate irregularly and becomes unreliable in continual changing environment. To address this limitation, we employ the MC Dropout technique \citep{gal2016dropout} on linear layers, enabling multiple forward propagations to obtain $m$ sets of probabilities for each sample. Subsequently, we calculate the uncertainty value $\mathcal{U}(x)$ for a given input $x$, which are formulated as: 
\begin{equation}
\mathcal{U} (x) =  \left( \frac{1}{m} \sum_{i=1}^m \|p_i(y|x) - \mu \|^2 \right) ^{\frac{1}{2}}
\label{eq:mc}
\end{equation}
Where $p_i(y|x)$ is the predicted probability of the input $x$ in the $i^{th}$ forward propagation and $\mu$ is the average value of $m$ times prediction.  
To dynamically adjust the scale factors ($\lambda_h$ and $\lambda_l$) based on the uncertainty score, the formulation is as follows:
\begin{equation}
\scriptstyle
   \left\{
    \begin{array}{l}
            \lambda_h = 1 + \mathcal{U} (x)\quad \lambda_l = 1 - \mathcal{U} (x), \quad \mathcal{U} (x) \ge \Theta\\  
            \lambda_h = 1 - \mathcal{U} (x)\quad \lambda_l = 1 + \mathcal{U} (x), \quad \mathcal{U} (x) < \Theta
        \end{array}
        \right.
\end{equation}
The threshold value of uncertainty is denoted as $\Theta$, where $\Theta = 0.2$. To realize the homeostasis of different domain knowledge, when facing the sample with a large uncertainty value, we adaptively increase the fusion weight of domain-specific knowledge ($\lambda_h$). Conversely, if the input has a low uncertainty value, the fusion weight of domain-shared knowledge ($\lambda_l$) will be increased.

\subsection{Optimization Objective}
Following previous CTTA work ~\citep{wang2022continual}, we leverage the teacher model $\mathcal{T}$ to generate the pseudo labels $\widetilde{y}$ for updating ViDAs. And the consistency loss $L_{ce}$ is the optimization objective.
\begin{equation}
 \mathcal{L}_{ce}(x) = - 
 \frac{1}{C}
 \sum_c^C \widetilde{y}(c) \log \hat{y}(c)
\label{eq:loss}
\end{equation}
Where $\hat{y}$ is the output of our student model $\mathcal{S}$ , $C$ means the number of categories. Same as previous works\citep{gan2023decorate}, we load the source pre-trained parameters to initialize the weight of both models and adopt the exponential moving average (EMA) to update the teacher model with ViDAs.
\begin{equation}
 \mathcal{T}^{t} = \alpha \mathcal{T}^{t-1} + (1-\alpha) \mathcal{S}^{t}
\label{eq:ema}
\end{equation}
Where $t$ is the time step. And we set the updating weight $\alpha = 0.999$ \citep{AnttiTarvainenetal2017}.

\section{Experiment}

In Section~\ref{sec:4.2} and ~\ref{sec:4.3}, we compare our method with other SOTA methods on classification and semantic segmentation CTTA.
In Section~\ref{sec:4.4}, we employ the foundation models (DINOv2 \citep{kirillov2023segment} and SAM \citep{oquab2023dinov2}) as the backbone and evaluate the efficacy of our method.
In Section~\ref{sec:4.5}, we further evaluate the domain generalization ability of the proposed method on unseen target domains.
Comprehensive ablation studies are conducted in Section~\ref{sec:4.6}. More quantitative comparisons and qualitative analyses are shown in the Appendix~\ref{ap:ex} and ~\ref{ap:qu}, respectively.

\subsection{Task settings and Datasets}
\label{sec:4.1}

\textbf{Dataset.} We evaluate our method on three classification CTTA benchmarks, including CIFAR10-to-CIFAR10C, CIFAR100-to-CIFAR100C \citep{krizhevsky2009learning} and ImageNet-to-ImageNet-C \citep{hendrycks2019benchmarking}. For segmentation CTTA \citep{yang2023exploring}, we evaluate our method on Cityscapes-to-ACDC, where the Cityscapes dataset \citep{cordts2016cityscapes} serves as the source domain, and the ACDC dataset \citep{sakaridis2021acdc} represents the target domains.

\noindent\textbf{CTTA Task setting.}
Following \citep{wang2022continual}, in classification CTTA tasks, we sequentially adapt the pre-trained source model to the fifteen target domains with the largest corruption severity (level 5). The online prediction results were evaluated immediately after encountering the input data.
Regarding segmentation CTTA \citep{yang2023exploring}, the source model is an off-the-shelf pre-trained on the Cityscapes dataset. As for the continual target domains, we utilize the ACDC dataset, which consists of images collected in four unseen visual conditions: Fog, Night, Rain, and Snow. To simulate continual environmental changes in real-life scenarios, we cyclically repeat the same sequence of target domains (Fog→Night→Rain→Snow) multiple times.

\noindent\textbf{Implementation Details.} 
In our CTTA experiments, we follow the implementation details specified in previous works \citep{wang2022continual} to ensure consistency and comparability. We adopt ViT-base \citep{dosovitskiy2020image} and ResNet \citep{he2016deep} as the backbone in the classification CTTA. In the case of ViT-base, we resize the input images to 224x224, while maintaining the original image resolution for other backbones. For segmentation CTTA, we adopt the pre-trained Segformer-B5 model~\citep{xie2021segformer} as the source model. We down-sample the input size from 1920x1080 to 960x540 for target domain data. The optimizer is performed using Adam~\citep{kingma2014adam} with $(\beta_1, \beta_2) = (0.9, 0.999)$. We set the learning rates to specific values for each task: 1e-4 for CIFAR10C, 5e-7 for ImageNetC, and 3e-4 for ACDC. To initialize our visual domain adapters, we train adapters for several iterations on classification datasets (e.g., ImageNet). We apply a range of image resolution scale factors [0.5, 0.75, 1.0, 1.25, 1.5, 1.75, 2.0] for the augmentation method and construct the teacher model inputs \citep{wang2022continual}. 

\subsection{The Effectiveness on Classification CTTA}
\label{sec:4.2}

\begin{table*}[htb]
\caption{\label{tab:imagenet}Classification error rate(\%) for ImageNet-to-ImageNet-C online CTTA task. Gain(\%) represents the percentage of improvement in model accuracy compared with the source method.}
\centering
\setlength\tabcolsep{2pt}
\begin{adjustbox}{width=1\linewidth,center=\linewidth}
\begin{tabular}{c|c|c|ccccccccccccccc|cc}
\toprule
Backbone & Method & REF &
 \rotatebox[origin=c]{70}{Gaussian} & \rotatebox[origin=c]{70}{shot} & \rotatebox[origin=c]{70}{impulse} & \rotatebox[origin=c]{70}{defocus} & \rotatebox[origin=c]{70}{glass} & \rotatebox[origin=c]{70}{motion} & \rotatebox[origin=c]{70}{zoom} & \rotatebox[origin=c]{70}{snow} & \rotatebox[origin=c]{70}{frost} & \rotatebox[origin=c]{70}{fog}  & \rotatebox[origin=c]{70}{brightness} & \rotatebox[origin=c]{70}{contrast} & \rotatebox[origin=c]{70}{elastic\_trans} & \rotatebox[origin=c]{70}{pixelate} & \rotatebox[origin=c]{70}{jpeg}
& Mean$\downarrow$ & Gain\\\midrule
\multirow{4}{*}{ResNet50}&Source  &\citep{KaimingHe2015DelvingDI}&97.8&97.1&98.2&81.7&89.8&85.2&78&83.5&77.1&75.9&41.3&94.5&82.5&79.3&68.6&82&0.0\\
&TENT & \citep{DequanWangetal2021} &81.6&74.6&72.7&77.6&73.8&65.5&55.3&61.6&63&51.7&38.2&72.1&50.8&47.4&53.3&62.6&+19.4\\
&CoTTA & \citep{wang2022continual}
&84.7 &82.1& 80.6& 81.3& 79.0& 68.6& 57.5& 60.3& 60.5& 48.3& 36.6& 66.1& 47.2& 41.2& 46.0&62.7&+19.3\\
&EcoTTA & \citep{song2023ecotta}
&-&-&-&-&-&-&-&-&-&-&-&-&-&-&-&63.4&+18.6\\
\rowcolor{green!10}&Ours & \textbf{Proposed} &\textbf{79.3}& \textbf{74.7}& \textbf{73.1}& \textbf{76.9}& \textbf{74.5}& \textbf{65.0}& \textbf{56.4}& \textbf{59.8}& \textbf{62.6}& \textbf{49.6}& \textbf{38.2}& \textbf{66.8}& \textbf{49.6}& \textbf{43.1}& \textbf{46.2}&\textbf{61.2}&\textbf{+20.8}\\
\hline
\multirow{6}{*}{ViT-base}&Source & \citep{dosovitskiy2020image}&53.0&51.8&52.1&68.5&78.8&58.5&63.3&49.9&54.2&57.7&26.4&91.4&57.5&38.0&36.2&55.8&0.0\\
&Pseudo & \citep{Leeetal2013}&45.2&40.4&41.6&51.3&53.9&45.6&47.7&40.4&45.7&93.8&98.5&99.9&99.9&98.9&99.6&61.2&-5.4\\
&TENT  &\citep{DequanWangetal2021} &52.2&48.9&49.2&65.8&73&54.5&58.4&44.0&47.7&50.3&23.9&72.8&55.7&34.4&33.9&51.0&+4.8\\
&CoTTA & \citep{wang2022continual}&52.9&51.6&51.4&68.3&78.1&57.1&62.0&48.2&52.7&55.3&25.9&90.0&56.4&36.4&35.2&54.8&+1.0\\
&VDP & \citep{gan2023decorate} &52.7&51.6&50.1&58.1&70.2&56.1&58.1&42.1&46.1&45.8&23.6&70.4&54.9&34.5&36.1&50.0&+5.8\\
\rowcolor{green!10}& \textbf{Ours} & \textbf{Proposed} & \textbf{47.7}& \textbf{42.5}& \textbf{42.9}& \textbf{52.2}& \textbf{56.9}& \textbf{45.5}& \textbf{48.9}& \textbf{38.9}& \textbf{42.7}& \textbf{40.7}& \textbf{24.3}& \textbf{52.8}& \textbf{49.1}& \textbf{33.5}& \textbf{33.1}& \textbf{43.4}& +\textbf{12.4}\\
\bottomrule
\end{tabular}
\end{adjustbox}
\vspace{-0.2cm}
\end{table*}

\noindent\textbf{ImageNet-to-ImageNet-C.} Given the source model pre-trained on ImageNet, we conduct CTTA on ImageNet-C, which consists of fifteen corruption types that occur sequentially during the test time. 
In Table \ref{tab:imagenet}, methods that utilize the ViT backbone achieve lower classification errors compared to those using the ResNet50 backbone, demonstrating ViT's superior generalization capability in the continually changing environment.
For ViT-base, the average classification error is up to 55.8\% when we directly test the source model on target domains.
Our method can outperform all previous methods, achieving a 12.4\% and 6.6\% improvement over the source model and previous SOTA method, respectively. Moreover, our method showcases remarkable performance across the majority of corruption types, highlighting its effective mitigation of error accumulation and catastrophic forgetting.
In addition, we conduct a 10-round CTTA experiment in Appendix \ref{ap:moqa}, which repeat 10 rounds of 15 corruption sequences in ImageNet-C. The performance of our method consistently improves over time, demonstrating its enduring robustness in the long-term adaptation process.

\begin{table}[t]
\begin{center}
    \begin{minipage}{0.48\textwidth}
    \vspace{-0.2cm}
    \centering
        \setlength\tabcolsep{0.043cm}
        \caption{\label{tab:cifar} 
        Average error rate (\%) for the standard CIFAR10-to-CIAFAR10C and CIFAR100-to-CIAFAR100C CTTA. All results are evaluated on the ViT-base, and the fine-grained performances are shown in Appendix \ref{ap:fi}.}
        \small
        \begin{tabular}{c|c|cccccc}
        \toprule
        Target&Method & Source& Tent& CoTTA& VDP & \cellcolor{green!10}\textbf{Ours}\\\hline
        \multirow{2}{*}{Cifar10C}&Mean$\downarrow$& 28.2&23.5&24.6&24.1 &\cellcolor{green!10}\textbf{20.7}\\
        &Gain$\uparrow$&0.0 &+4.7&+3.6&+4.1 &\cellcolor{green!10}\textbf{+7.5}\\\hline
        \multirow{2}{*}{Cifar100C}&Mean$\downarrow$& 35.4&32.1&34.8&35.0 &\cellcolor{green!10}\textbf{27.3}\\
        &Gain$\uparrow$&0.0 &+3.3&+0.7&+0.4&\cellcolor{green!10}\textbf{+8.1}\\
        \bottomrule
        \end{tabular}
    \end{minipage}
    \hspace{0.02\textwidth}
    \begin{minipage}{0.48\textwidth}
    \vspace{-0.3cm}
        \centering
        \setlength\tabcolsep{0.07cm}
        \caption{\label{tab:CIFAR10-to-CIFAR10c-dino} Average error rate (\%) for the CIFAR10-to-CIFAR10C CTTA task. All results are evaluated on the ViT-Base, which uses the pre-trained encoder parameter of foundation large-scale models (DINOv2 and SAM).}
        \small
        \begin{tabular}{c|c|cccc}
        \toprule
            Backbone&Method & Source & Tent & CoTTA & \cellcolor{green!10}Ours\\\hline
          \multirow{2}{*}{DINOv2} & \justifying Mean$\downarrow$ & 25.0 & 21.7 & 29.3 &  \cellcolor{green!10} \textbf{20.2}\\
            
            &\justifying Gain$\uparrow$ & $0.0$ & $+3.2$ & $-4.3$ &   \cellcolor{green!10}\textbf{+4.8}\\\hline
           \multirow{2}{*}{SAM}  &Mean$\downarrow$ & 39.3 & 37.5 & 39.4 &   \cellcolor{green!10}\textbf{34.1}\\
            
            &Gain$\uparrow$ & $0.0$ & $+1.8$ & $-0.1$ &   \cellcolor{green!10}\textbf{+5.2}\\
            \bottomrule
        \end{tabular}
    \end{minipage}
\end{center}
\vspace{-1em}
\end{table}

To further validate the effectiveness of our method, we conduct experiments on \textbf{CIFAR10-to-CIFAR10C} and \textbf{CIFAR100-to-CIFAR100C}. As illustrated in Table \ref{tab:cifar}, in CIFAR10C, our approach achieved a 2.8\% improvement compared to the previous SOTA model. We extend our evaluation to CIFAR100C, which comprises a larger number of categories in each domain. Our approach surpasses all previous methods, which show the same trend as the above CTTA experiments. Therefore, the results prove that our method mitigates the challenges posed by continual distribution shifts, regardless of the number of categories present in each domain. 
In addition, we provide supplementary CTTA experiments utilizing convolutional backbones in the Appendix \ref{ap:conv}.

\subsection{The Effectiveness on Segmentation CTTA}
\label{sec:4.3}

\begin{table*}[htb]
\vspace{-0.1cm}
\caption{\label{tab:ACDC} \textbf{Performance comparison for Cityscape-to-ACDC CTTA.} We sequentially repeat the same sequence of target domains three times. Mean is the average score of mIoU.}
\centering
\setlength\tabcolsep{2pt}
\begin{adjustbox}{width=1\linewidth,center=\linewidth}
\begin{tabular}{c|c|ccccc|ccccc|ccccc|c|c }
\hline

\multicolumn{2}{c|}{Time}     & \multicolumn{15}{c}{$t$ \makebox[10cm]{\rightarrowfill} }                                                                              \\ \hline
\multicolumn{2}{c|}{Round}          & \multicolumn{5}{c|}{1}    & \multicolumn{5}{c|}{2}     & \multicolumn{5}{c|}{3}  & \multirow{2}{*}{Mean$\uparrow$}   & \multirow{2}{*}{Gain}  \\ \cline{1-17}
Method & REF & Fog & Night & Rain & Snow & Mean$\uparrow$ & Fog & Night & Rain & Snow  & Mean$\uparrow$ & Fog & Night & Rain & Snow & Mean$\uparrow$ & \\ \hline
Source  & \citep{xie2021segformer}   &69.1&40.3&59.7&57.8&56.7&69.1&40.3&59.7& 	57.8&56.7&69.1&40.3&59.7& 57.8&56.7&56.7&/\\ 
TENT   & \citep{wang2020tent} &69.0&40.2&60.1&57.3&56.7&68.3&39.0&60.1& 	56.3&55.9&67.5&37.8&59.6&55.0&55.0&55.7&-1.0\\ 
CoTTA  & \citep{wang2022continual}  &70.9&41.2&62.4&59.7&58.6&70.9&41.1&62.6& 	59.7&58.6&70.9&41.0&62.7&59.7&58.6&58.6&+1.9\\ 
DePT   & \citep{gao2022visual}
&71.0&40.8&58.2&56.8&56.5&68.2&40.0&55.4&53.7& 54.3&66.4&38.0&47.3&47.2&49.7&53.4&-3.3\\
VDP  & \citep{gan2023decorate}  &70.5&41.1&62.1&59.5&  58.3    &70.4&41.1&62.2&59.4& 58.2     & 70.4&41.0&62.2&59.4& 58.2   &  58.2 & +1.5\\
 \rowcolor{green!10}\textbf{Ours} & \textbf{Proposed} & \textbf{71.6}& \textbf{43.2}& \textbf{66.0}& \textbf{63.4}& \textbf{61.1}& 
  \textbf{73.2}& \textbf{44.5}& \textbf{67.0}& \textbf{63.9}& \textbf{62.2} 
 & \textbf{73.2}& \textbf{44.6}& \textbf{67.2}& \textbf{64.2}& \textbf{62.3}      & \textbf{61.9} 
 & +\textbf{5.2}\\\bottomrule
 \hline
\end{tabular}
\end{adjustbox}
\vspace{-0.2cm}
\label{tab:CTTA}
\end{table*}

\noindent\textbf{Cityscapes-to-ACDC.}
As presented in Table \ref{tab:ACDC}, we observed a gradual decrease in the mIoUs of TENT and DePT over time, indicating the occurrence of catastrophic forgetting. In contrast, our method has a continual improvement of average mIoU (61.1→62.2→62.3) when the same sequence of target domains is repeated. Significantly, the proposed method surpasses the previous SOTA CTTA method \citep{wang2022continual} by achieving a 3.3\% increase in mIoU. This notable improvement showcases our method's ability to adapt continuously to dynamic target domains in the pixel-level task. The 10 rounds semantic segmentation CTTA experiments are shown in Appendix \ref{ap:seg}.

\subsection{Continual Adapting for Foundation Models}
\label{sec:4.4}

Foundation models~\citep{bommasani2021opportunities} are trained on large-scale datasets, endowing them with powerful generalization capabilities and the ability to capture representations of common features. However, performing full fine-tuning on the foundation model is time-consuming and economically impractical. Hence, our adaptation method proves valuable by enhancing the continual transfer performance of foundation models. As indicated in Table ~\ref{tab:CIFAR10-to-CIFAR10c-dino}, we introduce foundation models as the pre-trained model and adapt them to continual target domains (CIFAR10C). Our approach achieved a performance improvement of 4.8\% on the representative image-level foundation model DINOv2~\citep{oquab2023dinov2} and 5.2\% on pixel-level foundation model SAM~\citep{kirillov2023segment}. Our method consistently and reliably improves the performance of the foundation model on the continually changing environment. Note that, we only use the pre-trained encoder of SAM and add a classification head, which is fine-tuned on the source domain. 
Our approach empowers the large-scale model with the capability of continuous transfer learning, without undermining its plasticity.
Additional CTTA experiments of foundation models are shown in Appendix \ref{ap:clafo} and \ref{ap:segfo}

\subsection{Domain Generalization on Unseen Continual Domains}
\label{sec:4.5}

\begin{table}[t]
\begin{center}
    \begin{minipage}{0.48\textwidth}
    \vspace{-0.1cm}
    \centering
        \setlength\tabcolsep{0.05cm}
        \caption{\label{tab:dg} The domain generalization comparisons on ImageNet-C. Results are evaluated on ViT-base. Mean and Gain(\%) represent the performance on unseen target domains.}
        \small
        \begin{tabular}{l|ccccc|c}
        \toprule
         &  \multicolumn{5}{c|}{\textbf{Directly test on unseen domains}}& \multicolumn{1}{c}{\textbf{Unseen}} \\ \hline
        Method & bri. & contrast & elastic & pixelate & jpeg 
        & Mean$\downarrow$ \\
        \hline
        Source&26.4&91.4&57.5&38.0&36.2&49.9\\
        Tent &25.8&91.9&57.0&37.2&35.7&49.5\\
       CoTTA &25.3&88.1&55.7&36.4&34.6&48.0\\
       \rowcolor{green!10} \textbf{Ours} & \textbf{24.6}& \textbf{68.2}& \textbf{49.8}& \textbf{34.7}& \textbf{34.1}& \textbf{42.3}\\ 
        \hline
        \end{tabular}
            \end{minipage}
    \hspace{0.02\textwidth}
    \begin{minipage}{0.48\textwidth}
    \vspace{-0.3cm}
     \centering
        \setlength\tabcolsep{0.1cm}
        \caption{\label{tab:ablation} 
         Average error rate (\%) for the ImageNet-to-ImageNet-C. $ViDA_h$ and $ViDA_l$ represent the high-rank and low-rank ViDAs. IHKA means inversed HKA strategy.}
        \small
        \begin{tabular}{c|cccc|c}
        \hline
         & $ViDA_h$& $ViDA_l$  & HKA & IHKA & Mean$\downarrow$ \\\hline
        $Ex_{1}$ & - & - & -& - &55.8 \\ 
        $Ex_{2}$& \checkmark & - & -& - & 50.7\\
        $Ex_{3}$ & -& \checkmark  &- & -& 51.2\\
        $Ex_{4}$  & \checkmark & \checkmark & -& - &45.6\\
        $Ex_{5}$  & \checkmark & \checkmark &\checkmark& - & 43.4\\
        $Ex_{6}$  & \checkmark & \checkmark &-& \checkmark & 46.3\\
        \hline
        \end{tabular}
    \end{minipage}
\end{center}
\vspace{-0.3cm}
\end{table}

To investigate the domain generalization (DG) ability of our method, we follow the leave-one-domain-out rule \citep{zhou2021domain, li2017deeper} to leverage 10/15 domains of ImageNet-C as source domains for model training while the rest (5/15 domains) are treated as target domains without any form of adaptation. Specifically, we first use our proposed method to continually adapt the source pre-trained model to 10/15 domains of ImageNet-C without any supervision. Then we directly test on the 5/15 unseen domains. Surprisingly, our method reduces 7.6\% on the average error on unseen domains (Table \ref{tab:dg}), which has a significant improvement over other methods. The promising results demonstrate that our method possesses DG ability by effectively extracting domain-shared knowledge. 
More DG experiments are provided in the supplementary Appendix \ref{ap:dg}.

\subsection{Ablation study}
\label{sec:4.6}

\textbf{Effectiveness of each component.} 
We conduct the ablation study on ImageNet-to-ImageNet-C CTTA scenario and evaluate the contribution of each component in our method, including high-rank ViDA ($ViDA_h$), low-rank ViDA ($ViDA_l$), and Homeostatic Knowledge Allotment (HKA) strategy.
As shown in Table \ref{tab:ablation} (Ex$_2$), by introducing the high-rank ViDA, the error decreases by 5.1\% compared to Ex$_1$, demonstrating that high-rank features can extract more domain-specific knowledge for adaptation in target domains.
As shown in $Ex_{3}$, low-rank ViDA gains 4.6\% improvement compared to $Ex_{1}$. The result proves that the domain-share knowledge extracted from low-rank feature can also improve the classification ability on continual target domains. $Ex_{4}$ has a remarkable improvement of 10.2\% overall, demonstrating that the two types of ViDA can compensate for each other in the continual adaptation process. $Ex_{5}$ achieves a 12.4\% improvement, demonstrating the effectiveness of the HKA strategy in enhancing the distinct domain representations of each type of ViDA. To further assess the effectiveness of HKA, we perform an additional experiment, denoted as $Ex_{6}$, by inverting the scale factors within the HKA strategy. 
Specifically, for samples exhibiting high uncertainty, we reduced $\lambda_h$ while increase $\lambda_l$.
This results in a marginal increase of 0.7\% error compared to $Ex_{4}$ and 2.9\% error compared to $Ex_{5}$. The additional ablation studies are shown in Appendix \ref{ap:ab}.

\section{Conclusion}

In this paper, we propose a homeostatic Visual Domain Adapter (ViDA) to address error accumulation and catastrophic forgetting problems in Continual Test-Time Adaptation (CTTA) tasks. And we investigate that the low-rank ViDA can disregard the impact of dynamic distribution shifts and prioritize the
extraction of domain-shared knowledge, and the high-rank ViDA can extract more reliable domain-specific knowledge. Meanwhile, we further propose a Homeostatic Knowledge Allotment (HKA) strategy to dynamically fuse the knowledge from low-rank and high-rank ViDAs, thus enhancing ViDAs’ distinct domain representations.

\textbf{Acknowledgements.} Shanghang Zhang is supported by the National Science and Technology Major Project of China (No. 2022ZD0117801). We thank Xiaoqi Li for the insightful discussions and for helping with writing.

\bibliography{iclr2024_conference}

\begin{thebibliography}{70}
\providecommand{\natexlab}[1]{#1}
\providecommand{\url}[1]{\texttt{#1}}
\expandafter\ifx\csname urlstyle\endcsname\relax
  \providecommand{\doi}[1]{doi: #1}\else
  \providecommand{\doi}{doi: \begingroup \urlstyle{rm}\Url}\fi

\bibitem[Allaway et~al.(2021)Allaway, Srikanth, and McKeown]{allaway2021adversarial}
Emily Allaway, Malavika Srikanth, and Kathleen McKeown.
\newblock Adversarial learning for zero-shot stance detection on social media.
\newblock \emph{arXiv preprint arXiv:2105.06603}, 2021.

\bibitem[Arnold et~al.(2019)Arnold, Al-Jarrah, Dianati, Fallah, Oxtoby, and Mouzakitis]{arnold2019survey}
Eduardo Arnold, Omar~Y Al-Jarrah, Mehrdad Dianati, Saber Fallah, David Oxtoby, and Alex Mouzakitis.
\newblock A survey on 3d object detection methods for autonomous driving applications.
\newblock \emph{IEEE Transactions on Intelligent Transportation Systems}, 20\penalty0 (10):\penalty0 3782--3795, 2019.

\bibitem[Bahng et~al.(2022)Bahng, Jahanian, Sankaranarayanan, and Isola]{Bahngetal2022}
Hyojin Bahng, Ali Jahanian, Swami Sankaranarayanan, and Phillip Isola.
\newblock Exploring visual prompts for adapting large-scale models.
\newblock 2022.

\bibitem[Ben-David et~al.(2006)Ben-David, Blitzer, Crammer, and Pereira]{ben2006analysis}
Shai Ben-David, John Blitzer, Koby Crammer, and Fernando Pereira.
\newblock Analysis of representations for domain adaptation.
\newblock \emph{Advances in neural information processing systems}, 19, 2006.

\bibitem[Ben-David et~al.(2010)Ben-David, Blitzer, Crammer, Kulesza, Pereira, and Vaughan]{ben2010theory}
Shai Ben-David, John Blitzer, Koby Crammer, Alex Kulesza, Fernando Pereira, and Jennifer~Wortman Vaughan.
\newblock A theory of learning from different domains.
\newblock \emph{Machine learning}, 79\penalty0 (1):\penalty0 151--175, 2010.

\bibitem[Bommasani et~al.(2021)Bommasani, Hudson, Adeli, Altman, Arora, von Arx, Bernstein, Bohg, Bosselut, Brunskill, et~al.]{bommasani2021opportunities}
Rishi Bommasani, Drew~A Hudson, Ehsan Adeli, Russ Altman, Simran Arora, Sydney von Arx, Michael~S Bernstein, Jeannette Bohg, Antoine Bosselut, Emma Brunskill, et~al.
\newblock On the opportunities and risks of foundation models.
\newblock \emph{arXiv preprint arXiv:2108.07258}, 2021.

\bibitem[Boudiaf et~al.(2023)Boudiaf, Denton, van Merriënboer, Dumoulin, and Triantafillou]{Boudiafetal2023}
Malik Boudiaf, Tom Denton, Bart van Merriënboer, Vincent Dumoulin, and Eleni Triantafillou.
\newblock In search for a generalizable method for source free domain adaptation.
\newblock 2023.

\bibitem[Chakrabarty et~al.(2023)Chakrabarty, Sreenivas, and Biswas]{chakrabarty2023sata}
Goirik Chakrabarty, Manogna Sreenivas, and Soma Biswas.
\newblock Sata: Source anchoring and target alignment network for continual test time adaptation.
\newblock \emph{arXiv preprint arXiv:2304.10113}, 2023.

\bibitem[Chen et~al.(2022)Chen, Wang, Darrell, and Ebrahimi]{Chenetal2022}
Dian Chen, Dequan Wang, Trevor Darrell, and Sayna Ebrahimi.
\newblock Contrastive test-time adaptation.
\newblock \emph{ArXiv}, abs/2204.10377, 2022.

\bibitem[Chen et~al.()Chen, Chongjian, Tong, Wang, Song, Wang, and Luo]{chen2022adaptformer}
Shoufa Chen, GE~Chongjian, Zhan Tong, Jiangliu Wang, Yibing Song, Jue Wang, and Ping Luo.
\newblock Adaptformer: Adapting vision transformers for scalable visual recognition.
\newblock In \emph{Advances in Neural Information Processing Systems}.

\bibitem[Cordts et~al.(2016)Cordts, Omran, Ramos, Rehfeld, Enzweiler, Benenson, Franke, Roth, and Schiele]{cordts2016cityscapes}
Marius Cordts, Mohamed Omran, Sebastian Ramos, Timo Rehfeld, Markus Enzweiler, Rodrigo Benenson, Uwe Franke, Stefan Roth, and Bernt Schiele.
\newblock The cityscapes dataset for semantic urban scene understanding.
\newblock In \emph{Proceedings of the IEEE conference on computer vision and pattern recognition}, pp.\  3213--3223, 2016.

\bibitem[Ding et~al.(2021)Ding, Zhang, Ma, Han, Ding, and Sun]{ding2021repvgg}
Xiaohan Ding, Xiangyu Zhang, Ningning Ma, Jungong Han, Guiguang Ding, and Jian Sun.
\newblock Repvgg: Making vgg-style convnets great again.
\newblock In \emph{Proceedings of the IEEE/CVF conference on computer vision and pattern recognition}, pp.\  13733--13742, 2021.

\bibitem[D{\"o}bler et~al.(2023)D{\"o}bler, Marsden, and Yang]{dobler2023robust}
Mario D{\"o}bler, Robert~A Marsden, and Bin Yang.
\newblock Robust mean teacher for continual and gradual test-time adaptation.
\newblock In \emph{Proceedings of the IEEE/CVF Conference on Computer Vision and Pattern Recognition}, pp.\  7704--7714, 2023.

\bibitem[Dosovitskiy et~al.(2020)Dosovitskiy, Beyer, Kolesnikov, Weissenborn, Zhai, Unterthiner, Dehghani, Minderer, Heigold, Gelly, et~al.]{dosovitskiy2020image}
Alexey Dosovitskiy, Lucas Beyer, Alexander Kolesnikov, Dirk Weissenborn, Xiaohua Zhai, Thomas Unterthiner, Mostafa Dehghani, Matthias Minderer, Georg Heigold, Sylvain Gelly, et~al.
\newblock An image is worth 16x16 words: Transformers for image recognition at scale.
\newblock \emph{arXiv preprint arXiv:2010.11929}, 2020.

\bibitem[Gal \& Ghahramani(2016)Gal and Ghahramani]{gal2016dropout}
Yarin Gal and Zoubin Ghahramani.
\newblock Dropout as a bayesian approximation: Representing model uncertainty in deep learning.
\newblock In \emph{international conference on machine learning}, pp.\  1050--1059. PMLR, 2016.

\bibitem[Gan et~al.(2022)Gan, Ma, Lou, Bai, Zhang, Shi, and Luo]{gan2022decorate}
Yulu Gan, Xianzheng Ma, Yihang Lou, Yan Bai, Renrui Zhang, Nian Shi, and Lin Luo.
\newblock Decorate the newcomers: Visual domain prompt for continual test time adaptation.
\newblock \emph{arXiv preprint arXiv:2212.04145}, 2022.

\bibitem[Gan et~al.(2023)Gan, Bai, Lou, Ma, Zhang, Shi, and Luo]{gan2023decorate}
Yulu Gan, Yan Bai, Yihang Lou, Xianzheng Ma, Renrui Zhang, Nian Shi, and Lin Luo.
\newblock Decorate the newcomers: Visual domain prompt for continual test time adaptation.
\newblock In \emph{Proceedings of the AAAI Conference on Artificial Intelligence}, volume~37, pp.\  7595--7603, 2023.

\bibitem[Ganin et~al.(2016)Ganin, Ustinova, Ajakan, Germain, Larochelle, Laviolette, Marchand, and Lempitsky]{ganin2016domain}
Yaroslav Ganin, Evgeniya Ustinova, Hana Ajakan, Pascal Germain, Hugo Larochelle, Fran{\c{c}}ois Laviolette, Mario Marchand, and Victor Lempitsky.
\newblock Domain-adversarial training of neural networks.
\newblock \emph{The journal of machine learning research}, 17\penalty0 (1):\penalty0 2096--2030, 2016.

\bibitem[Gao et~al.(2021)Gao, Fisch, and Chen]{gao2021making}
Tianyu Gao, Adam Fisch, and Danqi Chen.
\newblock Making pre-trained language models better few-shot learners.
\newblock In \emph{Joint Conference of the 59th Annual Meeting of the Association for Computational Linguistics and the 11th International Joint Conference on Natural Language Processing, ACL-IJCNLP 2021}, pp.\  3816--3830. Association for Computational Linguistics (ACL), 2021.

\bibitem[Gao et~al.(2022)Gao, Shi, Zhu, Wang, Tang, Zhou, Li, and Metaxas]{gao2022visual}
Yunhe Gao, Xingjian Shi, Yi~Zhu, Hao Wang, Zhiqiang Tang, Xiong Zhou, Mu~Li, and Dimitris~N Metaxas.
\newblock Visual prompt tuning for test-time domain adaptation.
\newblock \emph{arXiv preprint arXiv:2210.04831}, 2022.

\bibitem[Gong et~al.(2022)Gong, Jeong, Kim, Kim, Shin, and Lee]{gong2022note}
Taesik Gong, Jongheon Jeong, Taewon Kim, Yewon Kim, Jinwoo Shin, and Sung-Ju Lee.
\newblock Note: Robust continual test-time adaptation against temporal correlation.
\newblock \emph{Advances in Neural Information Processing Systems}, 35:\penalty0 27253--27266, 2022.

\bibitem[He et~al.(2015)He, Zhang, Ren, and Sun]{KaimingHe2015DelvingDI}
Kaiming He, Xiangyu Zhang, Shaoqing Ren, and Jian Sun.
\newblock Delving deep into rectifiers: Surpassing human-level performance on imagenet classification.
\newblock \emph{international conference on computer vision}, 2015.

\bibitem[He et~al.(2016)He, Zhang, Ren, and Sun]{he2016deep}
Kaiming He, Xiangyu Zhang, Shaoqing Ren, and Jian Sun.
\newblock Deep residual learning for image recognition.
\newblock In \emph{Proceedings of the IEEE conference on computer vision and pattern recognition}, pp.\  770--778, 2016.

\bibitem[He et~al.(2021)He, Liu, Ye, Tan, Ding, Cheng, Low, Bing, and Si]{he2021effectiveness}
Ruidan He, Linlin Liu, Hai Ye, Qingyu Tan, Bosheng Ding, Liying Cheng, Jia-Wei Low, Lidong Bing, and Luo Si.
\newblock On the effectiveness of adapter-based tuning for pretrained language model adaptation.
\newblock \emph{arXiv preprint arXiv:2106.03164}, 2021.

\bibitem[Hendrycks \& Dietterich(2019)Hendrycks and Dietterich]{hendrycks2019benchmarking}
Dan Hendrycks and Thomas Dietterich.
\newblock Benchmarking neural network robustness to common corruptions and perturbations.
\newblock \emph{arXiv preprint arXiv:1903.12261}, 2019.

\bibitem[Houlsby et~al.(2019)Houlsby, Giurgiu, Jastrzebski, Morrone, De~Laroussilhe, Gesmundo, Attariyan, and Gelly]{houlsby2019parameter}
Neil Houlsby, Andrei Giurgiu, Stanislaw Jastrzebski, Bruna Morrone, Quentin De~Laroussilhe, Andrea Gesmundo, Mona Attariyan, and Sylvain Gelly.
\newblock Parameter-efficient transfer learning for nlp.
\newblock In \emph{International Conference on Machine Learning}, pp.\  2790--2799. PMLR, 2019.

\bibitem[Hu et~al.(2021)Hu, Shen, Wallis, Allen-Zhu, Li, Wang, Wang, and Chen]{hu2021lora}
Edward~J Hu, Yelong Shen, Phillip Wallis, Zeyuan Allen-Zhu, Yuanzhi Li, Shean Wang, Lu~Wang, and Weizhu Chen.
\newblock Lora: Low-rank adaptation of large language models.
\newblock \emph{arXiv preprint arXiv:2106.09685}, 2021.

\bibitem[Hu et~al.(2022)Hu, Ding, Wang, Liu, Wang, Li, Wu, and Sun]{hu2022knowledgeable}
Shengding Hu, Ning Ding, Huadong Wang, Zhiyuan Liu, Jingang Wang, Juanzi Li, Wei Wu, and Maosong Sun.
\newblock Knowledgeable prompt-tuning: Incorporating knowledge into prompt verbalizer for text classification.
\newblock In \emph{Proceedings of the 60th Annual Meeting of the Association for Computational Linguistics (Volume 1: Long Papers)}, pp.\  2225--2240, 2022.

\bibitem[Kingma \& Ba(2014)Kingma and Ba]{kingma2014adam}
Diederik~P Kingma and Jimmy Ba.
\newblock Adam: A method for stochastic optimization.
\newblock \emph{arXiv preprint arXiv:1412.6980}, 2014.

\bibitem[Kirillov et~al.(2023)Kirillov, Mintun, Ravi, Mao, Rolland, Gustafson, Xiao, Whitehead, Berg, Lo, et~al.]{kirillov2023segment}
Alexander Kirillov, Eric Mintun, Nikhila Ravi, Hanzi Mao, Chloe Rolland, Laura Gustafson, Tete Xiao, Spencer Whitehead, Alexander~C Berg, Wan-Yen Lo, et~al.
\newblock Segment anything.
\newblock \emph{arXiv preprint arXiv:2304.02643}, 2023.

\bibitem[Krizhevsky et~al.(2009)Krizhevsky, Hinton, et~al.]{krizhevsky2009learning}
Alex Krizhevsky, Geoffrey Hinton, et~al.
\newblock Learning multiple layers of features from tiny images.
\newblock 2009.

\bibitem[Kundu et~al.(2020)Kundu, Venkat, M, and Babu]{Kunduetal2022}
Jogendra~Nath Kundu, Naveen Venkat, Rahul M, and R.~Venkatesh Babu.
\newblock Universal source-free domain adaptation.
\newblock 2020.

\bibitem[Lee(2013)]{Leeetal2013}
Dong-Hyun Lee.
\newblock Pseudo-label : The simple and efficient semi-supervised learning method for deep neural networks.
\newblock 2013.

\bibitem[Li et~al.(2017)Li, Yang, Song, and Hospedales]{li2017deeper}
Da~Li, Yongxin Yang, Yi-Zhe Song, and Timothy~M Hospedales.
\newblock Deeper, broader and artier domain generalization.
\newblock In \emph{Proceedings of the IEEE international conference on computer vision}, pp.\  5542--5550, 2017.

\bibitem[Li et~al.(2020)Li, Chen, Ding, Zhu, Lu, and Shen]{li2020maximum}
Jingjing Li, Erpeng Chen, Zhengming Ding, Lei Zhu, Ke~Lu, and Heng~Tao Shen.
\newblock Maximum density divergence for domain adaptation.
\newblock \emph{IEEE transactions on pattern analysis and machine intelligence}, 43\penalty0 (11):\penalty0 3918--3930, 2020.

\bibitem[Li et~al.(2023)Li, Zhang, Geng, Geng, Long, Shen, Zhang, Liu, and Dong]{li2023manipllm}
Xiaoqi Li, Mingxu Zhang, Yiran Geng, Haoran Geng, Yuxing Long, Yan Shen, Renrui Zhang, Jiaming Liu, and Hao Dong.
\newblock Manipllm: Embodied multimodal large language model for object-centric robotic manipulation.
\newblock \emph{arXiv preprint arXiv:2312.16217}, 2023.

\bibitem[Liang et~al.(2020)Liang, Hu, and Feng]{Liangetal2020}
Jian Liang, D.~Hu, and Jiashi Feng.
\newblock Do we really need to access the source data? source hypothesis transfer for unsupervised domain adaptation.
\newblock In \emph{ICML}, 2020.

\bibitem[Liang et~al.(2023)Liang, He, and Tan]{liang2023comprehensive}
Jian Liang, Ran He, and Tieniu Tan.
\newblock A comprehensive survey on test-time adaptation under distribution shifts.
\newblock \emph{arXiv preprint arXiv:2303.15361}, 2023.

\bibitem[MacQueen(1967)]{macqueen1967classification}
J~MacQueen.
\newblock Classification and analysis of multivariate observations.
\newblock In \emph{5th Berkeley Symp. Math. Statist. Probability}, pp.\  281--297. University of California Los Angeles LA USA, 1967.

\bibitem[Mummadi et~al.(2021)Mummadi, Hutmacher, Rambach, Levinkov, Brox, and Metzen]{mummadi2021test}
Chaithanya~Kumar Mummadi, Robin Hutmacher, Kilian Rambach, Evgeny Levinkov, Thomas Brox, and Jan~Hendrik Metzen.
\newblock Test-time adaptation to distribution shift by confidence maximization and input transformation.
\newblock \emph{arXiv preprint arXiv:2106.14999}, 2021.

\bibitem[Ni et~al.(2023)Ni, Yang, Liu, Li, Jiao, Xu, Chen, Liu, and Zhang]{ni2023distribution}
Jiayi Ni, Senqiao Yang, Jiaming Liu, Xiaoqi Li, Wenyu Jiao, Ran Xu, Zehui Chen, Yi~Liu, and Shanghang Zhang.
\newblock Distribution-aware continual test time adaptation for semantic segmentation.
\newblock \emph{arXiv preprint arXiv:2309.13604}, 2023.

\bibitem[Niu et~al.(2023)Niu, Wu, Zhang, Wen, Chen, Zhao, and Tan]{niu2023towards}
Shuaicheng Niu, Jiaxiang Wu, Yifan Zhang, Zhiquan Wen, Yaofo Chen, Peilin Zhao, and Mingkui Tan.
\newblock Towards stable test-time adaptation in dynamic wild world.
\newblock \emph{arXiv preprint arXiv:2302.12400}, 2023.

\bibitem[Oquab et~al.(2023)Oquab, Darcet, Moutakanni, Vo, Szafraniec, Khalidov, Fernandez, Haziza, Massa, El-Nouby, et~al.]{oquab2023dinov2}
Maxime Oquab, Timoth{\'e}e Darcet, Th{\'e}o Moutakanni, Huy Vo, Marc Szafraniec, Vasil Khalidov, Pierre Fernandez, Daniel Haziza, Francisco Massa, Alaaeldin El-Nouby, et~al.
\newblock Dinov2: Learning robust visual features without supervision.
\newblock \emph{arXiv preprint arXiv:2304.07193}, 2023.

\bibitem[Ovadia et~al.(2019)Ovadia, Fertig, Ren, Nado, Sculley, Nowozin, Dillon, Lakshminarayanan, and Snoek]{ovadia2019can}
Yaniv Ovadia, Emily Fertig, Jie Ren, Zachary Nado, David Sculley, Sebastian Nowozin, Joshua Dillon, Balaji Lakshminarayanan, and Jasper Snoek.
\newblock Can you trust your model's uncertainty? evaluating predictive uncertainty under dataset shift.
\newblock \emph{Advances in neural information processing systems}, 32, 2019.

\bibitem[Qin et~al.(2021)Qin, Wang, Su, Lin, Ding, Liu, Li, Hou, Li, Sun, et~al.]{qin2021exploring}
Yujia Qin, Xiaozhi Wang, Yusheng Su, Yankai Lin, Ning Ding, Zhiyuan Liu, Juanzi Li, Lei Hou, Peng Li, Maosong Sun, et~al.
\newblock Exploring low-dimensional intrinsic task subspace via prompt tuning.
\newblock \emph{arXiv preprint arXiv:2110.07867}, 2021.

\bibitem[Rebuffi et~al.(2017)Rebuffi, Bilen, and Vedaldi]{rebuffi2017learning}
Sylvestre-Alvise Rebuffi, Hakan Bilen, and Andrea Vedaldi.
\newblock Learning multiple visual domains with residual adapters.
\newblock \emph{Advances in neural information processing systems}, 30, 2017.

\bibitem[Rebuffi et~al.(2018)Rebuffi, Bilen, and Vedaldi]{rebuffi2018efficient}
Sylvestre-Alvise Rebuffi, Hakan Bilen, and Andrea Vedaldi.
\newblock Efficient parametrization of multi-domain deep neural networks.
\newblock In \emph{Proceedings of the IEEE Conference on Computer Vision and Pattern Recognition}, pp.\  8119--8127, 2018.

\bibitem[Roy et~al.(2022)Roy, Trapp, Pilzer, Kannala, Sebe, Ricci, and Solin]{roy2022uncertainty}
Subhankar Roy, Martin Trapp, Andrea Pilzer, Juho Kannala, Nicu Sebe, Elisa Ricci, and Arno Solin.
\newblock Uncertainty-guided source-free domain adaptation.
\newblock In \emph{Computer Vision--ECCV 2022: 17th European Conference, Tel Aviv, Israel, October 23--27, 2022, Proceedings, Part XXV}, pp.\  537--555. Springer, 2022.

\bibitem[Ruder \& Plank(2017)Ruder and Plank]{ruder2017learning}
Sebastian Ruder and Barbara Plank.
\newblock Learning to select data for transfer learning with bayesian optimization.
\newblock \emph{arXiv preprint arXiv:1707.05246}, 2017.

\bibitem[Sakaridis et~al.(2021)Sakaridis, Dai, and Van~Gool]{sakaridis2021acdc}
Christos Sakaridis, Dengxin Dai, and Luc Van~Gool.
\newblock Acdc: The adverse conditions dataset with correspondences for semantic driving scene understanding.
\newblock In \emph{Proceedings of the IEEE/CVF International Conference on Computer Vision}, pp.\  10765--10775, 2021.

\bibitem[Schneider et~al.(2020)Schneider, Rusak, Eck, Bringmann, Brendel, and Bethge]{Schneideretal2020}
Steffen Schneider, Evgenia Rusak, Luisa Eck, Oliver Bringmann, Wieland Brendel, and Matthias Bethge.
\newblock Improving robustness against common corruptions by covariate shift adaptation.
\newblock \emph{ArXiv}, abs/2006.16971, 2020.

\bibitem[Song et~al.(2023)Song, Lee, Kweon, and Choi]{song2023ecotta}
Junha Song, Jungsoo Lee, In~So Kweon, and Sungha Choi.
\newblock Ecotta: Memory-efficient continual test-time adaptation via self-distilled regularization.
\newblock In \emph{Proceedings of the IEEE/CVF Conference on Computer Vision and Pattern Recognition}, pp.\  11920--11929, 2023.

\bibitem[Sung et~al.(2022)Sung, Cho, and Bansal]{sung2022vl}
Yi-Lin Sung, Jaemin Cho, and Mohit Bansal.
\newblock Vl-adapter: Parameter-efficient transfer learning for vision-and-language tasks.
\newblock In \emph{Proceedings of the IEEE/CVF Conference on Computer Vision and Pattern Recognition}, pp.\  5227--5237, 2022.

\bibitem[Tarvainen \& Valpola(2017{\natexlab{a}})Tarvainen and Valpola]{AnttiTarvainenetal2017}
Antti Tarvainen and Harri Valpola.
\newblock Mean teachers are better role models: Weight-averaged consistency targets improve semi-supervised deep learning results.
\newblock \emph{Learning}, 2017{\natexlab{a}}.

\bibitem[Tarvainen \& Valpola(2017{\natexlab{b}})Tarvainen and Valpola]{tarvainen2017mean}
Antti Tarvainen and Harri Valpola.
\newblock Mean teachers are better role models: Weight-averaged consistency targets improve semi-supervised deep learning results.
\newblock \emph{Advances in neural information processing systems}, 30, 2017{\natexlab{b}}.

\bibitem[Van~der Maaten \& Hinton(2008)Van~der Maaten and Hinton]{van2008visualizing}
Laurens Van~der Maaten and Geoffrey Hinton.
\newblock Visualizing data using t-sne.
\newblock \emph{Journal of machine learning research}, 9\penalty0 (11), 2008.

\bibitem[Vu et~al.(2022)Vu, Lester, Constant, Al-Rfou, and Cer]{vu2022spot}
Tu~Vu, Brian Lester, Noah Constant, Rami Al-Rfou, and Daniel Cer.
\newblock Spot: Better frozen model adaptation through soft prompt transfer.
\newblock In \emph{Proceedings of the 60th Annual Meeting of the Association for Computational Linguistics (Volume 1: Long Papers)}, pp.\  5039--5059, 2022.

\bibitem[Wang et~al.(2020)Wang, Shelhamer, Liu, Olshausen, and Darrell]{wang2020tent}
Dequan Wang, Evan Shelhamer, Shaoteng Liu, Bruno Olshausen, and Trevor Darrell.
\newblock Tent: Fully test-time adaptation by entropy minimization.
\newblock \emph{arXiv preprint arXiv:2006.10726}, 2020.

\bibitem[Wang et~al.(2021)Wang, Shelhamer, Liu, Olshausen, and Darrell]{DequanWangetal2021}
Dequan Wang, Evan Shelhamer, Shaoteng Liu, Bruno~A. Olshausen, and Trevor Darrell.
\newblock Tent: Fully test-time adaptation by entropy minimization.
\newblock In \emph{ICLR}, 2021.

\bibitem[Wang et~al.(2022)Wang, Fink, Van~Gool, and Dai]{wang2022continual}
Qin Wang, Olga Fink, Luc Van~Gool, and Dengxin Dai.
\newblock Continual test-time domain adaptation.
\newblock In \emph{Proceedings of the IEEE/CVF Conference on Computer Vision and Pattern Recognition}, pp.\  7201--7211, 2022.

\bibitem[Xie et~al.(2021)Xie, Wang, Yu, Anandkumar, Alvarez, and Luo]{xie2021segformer}
Enze Xie, Wenhai Wang, Zhiding Yu, Anima Anandkumar, Jose~M Alvarez, and Ping Luo.
\newblock Segformer: Simple and efficient design for semantic segmentation with transformers.
\newblock \emph{Advances in Neural Information Processing Systems}, 34:\penalty0 12077--12090, 2021.

\bibitem[Yang et~al.(2023{\natexlab{a}})Yang, Liu, Zhang, Pan, Guo, Li, Chen, Gao, Guo, and Zhang]{yang2023lidar}
Senqiao Yang, Jiaming Liu, Ray Zhang, Mingjie Pan, Zoey Guo, Xiaoqi Li, Zehui Chen, Peng Gao, Yandong Guo, and Shanghang Zhang.
\newblock Lidar-llm: Exploring the potential of large language models for 3d lidar understanding.
\newblock \emph{arXiv preprint arXiv:2312.14074}, 2023{\natexlab{a}}.

\bibitem[Yang et~al.(2023{\natexlab{b}})Yang, Wu, Liu, Li, Zhang, Pan, and Zhang]{yang2023exploring}
Senqiao Yang, Jiarui Wu, Jiaming Liu, Xiaoqi Li, Qizhe Zhang, Mingjie Pan, and Shanghang Zhang.
\newblock Exploring sparse visual prompt for cross-domain semantic segmentation.
\newblock \emph{arXiv preprint arXiv:2303.09792}, 2023{\natexlab{b}}.

\bibitem[Yang et~al.(2021)Yang, Wang, van~de Weijer, Herranz, and Jui]{ShiqiYangetal2021}
Shiqi Yang, Yaxing Wang, Joost van~de Weijer, Luis Herranz, and Shangling Jui.
\newblock Generalized source-free domain adaptation.
\newblock \emph{international conference on computer vision}, 2021.

\bibitem[Yuan et~al.(2023)Yuan, Xie, and Li]{yuan2023robust}
Longhui Yuan, Binhui Xie, and Shuang Li.
\newblock Robust test-time adaptation in dynamic scenarios.
\newblock In \emph{Proceedings of the IEEE/CVF Conference on Computer Vision and Pattern Recognition}, pp.\  15922--15932, 2023.

\bibitem[Zagoruyko \& Komodakis(2016)Zagoruyko and Komodakis]{SergeyZagoruykoetal2016}
Sergey Zagoruyko and Nikos Komodakis.
\newblock Wide residual networks.
\newblock \emph{british machine vision conference}, 2016.

\bibitem[Zaken et~al.(2021)Zaken, Ravfogel, and Goldberg]{zaken2021bitfit}
Elad~Ben Zaken, Shauli Ravfogel, and Yoav Goldberg.
\newblock Bitfit: Simple parameter-efficient fine-tuning for transformer-based masked language-models.
\newblock \emph{arXiv preprint arXiv:2106.10199}, 2021.

\bibitem[Zheng et~al.(2021)Zheng, Lu, Zhao, Zhu, Luo, Wang, Fu, Feng, Xiang, Torr, et~al.]{zheng2021rethinking}
Sixiao Zheng, Jiachen Lu, Hengshuang Zhao, Xiatian Zhu, Zekun Luo, Yabiao Wang, Yanwei Fu, Jianfeng Feng, Tao Xiang, Philip~HS Torr, et~al.
\newblock Rethinking semantic segmentation from a sequence-to-sequence perspective with transformers.
\newblock In \emph{Proceedings of the IEEE/CVF conference on computer vision and pattern recognition}, pp.\  6881--6890, 2021.

\bibitem[Zhou et~al.(2021)Zhou, Liu, Qiao, Xiang, and Loy]{zhou2021domain}
Kaiyang Zhou, Ziwei Liu, Yu~Qiao, Tao Xiang, and Chen~Change Loy.
\newblock Domain generalization in vision: A survey.
\newblock \emph{arXiv preprint arXiv:2103.02503}, 2021.

\bibitem[Zhu et~al.(2020)Zhu, Su, Lu, Li, Wang, and Dai]{zhu2020deformable}
Xizhou Zhu, Weijie Su, Lewei Lu, Bin Li, Xiaogang Wang, and Jifeng Dai.
\newblock Deformable detr: Deformable transformers for end-to-end object detection.
\newblock \emph{arXiv preprint arXiv:2010.04159}, 2020.

\end{thebibliography}
\bibliographystyle{iclr2024_conference}

\clearpage

\appendix
\section{Appendix}

The supplementary materials presented in this paper offer a comprehensive quantitative and qualitative analysis of the proposed method. In Appendix~\ref{ap:mo}, we provide additional empirical observations and justifications for our motivation, including specifically designed quantitative analysis, qualitative analysis of the distribution, and justification for distribution divergence. Additionally, we present extra continual adaptation experiments for Foundation Models in Appendices~\ref{ap:clafo} and~\ref{ap:segfo}, which are conducted on ImageNet-to-ImageNet-C and Cityscape-to-ACDC scenarios. To assess the domain generalization ability of our method, we conducted additional experiments directly testing a varying number of unseen domains in Appendix~\ref{ap:dg}. The ablation study on middle-layer dimension is described in Appendix~\ref{ap:ab}. Furthermore, Appendix~\ref{ap:conv} presents additional CTTA classification experiments utilizing the convolutional backbone, while Appendix~\ref{ap:seg} outlines 10 rounds of semantic segmentation CTTA experiments. We provide an additional qualitative analysis in Appendix~\ref{ap:qu}. 
Moreover, we extend the classification results of our submission to include fine-grained performance in Appendix \ref{ap:fi}, showcasing the error rates across fifteen corruption types.

\section{Supplementary Justifications for Motivation}
\label{ap:mo}
The study of Continual Test-Time Adaptation (CTTA) poses significant challenges, particularly in addressing error accumulation and catastrophic forgetting \citep{wang2022continual,gan2023decorate}. Notably, the use of adapters with low-rank and high-rank features have demonstrated promising results in mitigating these challenges in our submission. In this section, we aim to provide comprehensive implementation details regarding the evidence supporting our motivation. 
Furthermore, we have introduced two new specially designed quantitative experiments in Section B.1. The first one is a 10-round CTTA experiment aimed at investigating the different domain representations of low-rank and high-rank ViDA during the long-term adaptation process. The second experiment explores the performance when all adapters adopt the same structures, such as using two high-rank adapters or two low-rank adapters. This experiment is conducted to validate that low-rank ViDA and high-rank ViDA complement each other in adapting to continually changing environments.

\subsection{Specially Designed Quantitative Analysis}
\label{ap:moqa}
To provide stronger evidence for our assumption, we have developed two evaluation approaches for both low-rank and high-rank adapters, which directly reflect their ability to extract domain-shared and domain-specific knowledge on ImageNet-to-ImageNet-C.

First, as shown in Figure \ref{apfig:mov} \textcolor{red}{(b)}, we execute a 10 rounds CTTA experiment on ImageNet-to-ImageNet-C. In this comprehensive experiment, we simulate a long-term adaptation scenario by repeating 10 rounds of 15 corruption sequences in the ImageNet-C. Remarkably, the high-rank ViDA achieves competitive results over other methods during the initial 1 to 3 rounds. This result demonstrates the high-rank feature's capacity to efficiently learn target domain-specific knowledge. However, an increment in error rates becomes obvious during the later rounds (rounds 5 to 10). The results validate the potential for encountering catastrophic forgetting when focusing exclusively on domain-specific knowledge. In contrast, the performance of the low-rank ViDA remains consistently robust throughout the continual adaptation process, verifying it concentrates more on extracting task-relevant knowledge and effectively prevents the catastrophic forgetting problem. And our proposed method consistently improves over time, demonstrating its robustness in the long-term adaptation process.

Second, we execute an ImageNet-to-ImageNet-C CTTA experiment using a combination of two high-rank adapters or two low-rank adapters, as shown in Table \ref{aptab:two}. To ensure fairness, we conducted these experiments without implementing the homeostatic knowledge allotment (HKA) strategy. Notably, the two low-rank adapters (Ex2) demonstrated consistently lower long-term error rates compared to the source model and two high-rank adapters. The above results can be attributed to the fact that the two low-rank ViDAs tend to learn general information and domain-shared knowledge during continual adaptation. However, our method outperforms the two low-rank adapters across 14 out of 15 corruption types. This indicates that solely relying on low-rank adapters without the involvement of high-rank adapters is insufficient to fit target domains and match their data distribution. On the other hand, the performance of the two high-rank adapters initially surpasses our approach (Ex4) in the early stages, covering the first few target domains. Nevertheless, a noticeable performance degradation becomes apparent in later target domains. This observation underscores a crucial finding: while increasing the number of high-rank ViDAs might enhance domain-specific knowledge acquisition during the initial phases of CTTA, it simultaneously exacerbates catastrophic forgetting throughout the entire adaptation process. In contrast, the fusion of both low-rank and high-rank ViDAs (Ex4) yields the most substantial improvement when compared to other configurations. Our collaborative approach leverages the distinct domain representations of these adapters to compensate for each other's advantages and achieve a more robust and effective continual adaptation.

\begin{table*}[t]
\caption{Classification error rate(\%) for ImageNet-to-ImageNet-C online CTTA task. Gain(\%) represents the percentage of improvement in model accuracy compared with the source method. 2$\times$ means using two same structures of adapters.}
\centering
\setlength\tabcolsep{2pt}
\begin{adjustbox}{width=1\linewidth,center=\linewidth}
\begin{tabular}{c|c|ccccccccccccccc|c}
\toprule
 &Method  &
 \rotatebox[origin=c]{70}{Gaussian} &  \rotatebox[origin=c]{70}{shot} &  \rotatebox[origin=c]{70}{impulse} &  \rotatebox[origin=c]{70}{defocus} &  \rotatebox[origin=c]{70}{glass} &  \rotatebox[origin=c]{70}{motion} &  \rotatebox[origin=c]{70}{zoom} &  \rotatebox[origin=c]{70}{snow} &  \rotatebox[origin=c]{70}{frost} &  \rotatebox[origin=c]{70}{fog}  &  \rotatebox[origin=c]{70}{brightness} &  \rotatebox[origin=c]{70}{contrast} &  \rotatebox[origin=c]{70}{elastic} &  \rotatebox[origin=c]{70}{pixelate} &  \rotatebox[origin=c]{70}{jpeg}
& Mean$\downarrow$ \\\midrule
Ex1&Source&53.0&51.8&52.1&68.5&78.8&58.5&63.3&49.9&54.2&57.7&26.4&91.4&57.5&38.0&36.2&55.8\\
Ex2& 2$\times$Low-rank  &51.2&48.3&47.8&56.9&66.5&49.3&54.4&42.1&47.0&45.2&23.2&65.6&52.0&\textbf{33.4}&33.5&47.7\\
Ex3& 2$\times$High-rank   &\textbf{50.1}&47.9&\textbf{45.3}&\textbf{54.8}&66.7&51.4&56.1&44.0&49.2&48.3&25.7&69.7&56.3&34.6&33.7&48.9\\
Ex4& Ours &50.3&\textbf{45.9}&45.5&55.1&\textbf{62.3}&\textbf{46.6}&\textbf{51.7}&\textbf{39.7}&\textbf{44.0}&\textbf{42.2}&\textbf{23.0}&\textbf{62.4}&\textbf{50.1}&33.4&\textbf{32.5}&\textbf{45.6}\\
\bottomrule
\end{tabular}
\end{adjustbox}
\label{aptab:two}
\end{table*}

\subsection{Additional Distribution Qualitative Analysis}
\begin{figure*}[t]
    \begin{center}
        \includegraphics[width=0.99\linewidth]{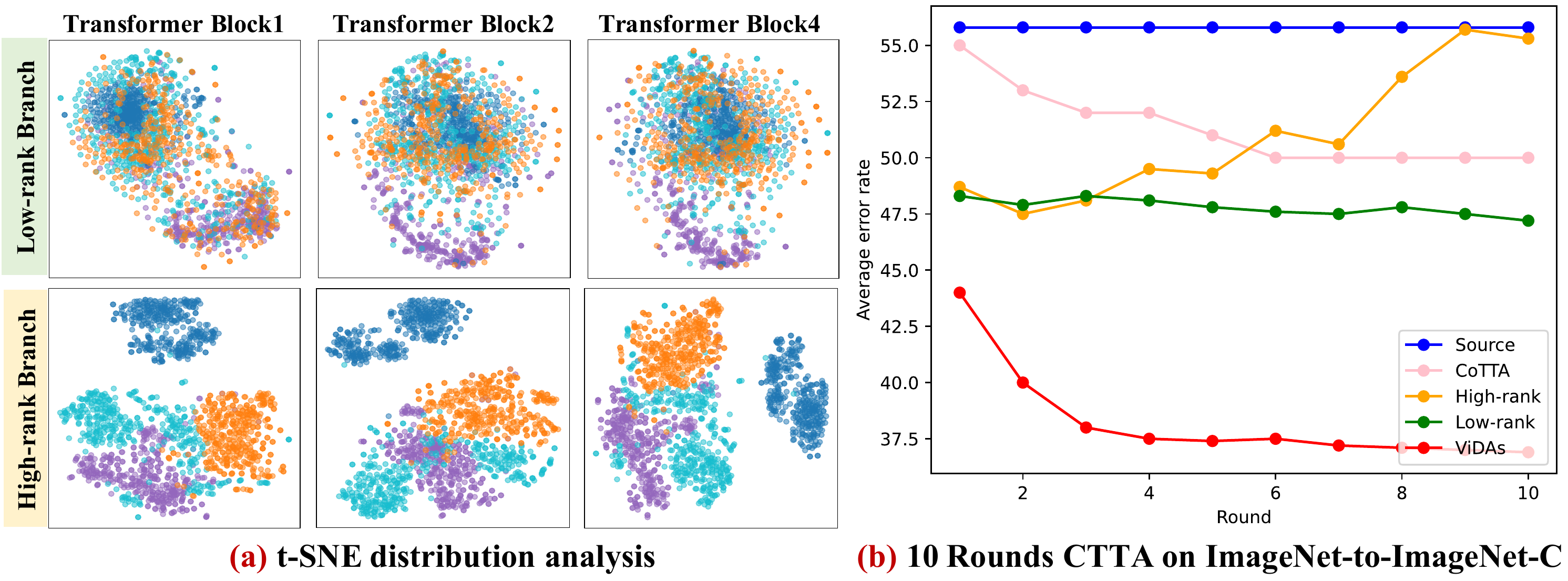}
    \end{center}
    \vspace{-0.2cm}
    \caption{\textcolor{red}{(a)} We conduct more t-SNE results for the low-rank adapter and high-rank adapter on the ACDC dataset. The first to third columns illustrate the feature distributions of transformer blocks 1, 2, and 4, respectively. \textcolor{red}{(b)} The 10 rounds CTTA experiment on ImageNet-to-ImageNet-C, repeating 10 rounds of 15 corruption sequences.
    }
    \label{apfig:mov}
\end{figure*}

We employed t-distributed stochastic neighbor embedding (t-SNE) \citep{van2008visualizing} to visualize the distribution of adapters across four continual target domains. This visualization was specifically conducted in the context of the Cityscapes-to-ACDC experiment, representing a scenario with continually changing real-world environments.In our submission, we perform t-SNE analysis on the outputs of the third transformer block in the Segformer-B5 model \citep{xie2021segformer}. The objective was to qualitatively compare the feature distributions of ViDAs with different dimension features. 
Furthermore, our findings revealed that the qualitative results obtained from different layers (i.e., transformer block 1, 2, and 4) of the Segformer-B5 model exhibited similar distribution representations.
As illustrated in Figure \ref{apfig:mov} \textcolor{red}{(a)}, there is a noticeable distribution gap due to the significant domain shift between the night domain and other domains. Interestingly, the low-rank ViDA effectively reduces the distribution distance across different target domains, indicating its focus on extracting task-relevant knowledge. On the other hand, the high-rank ViDA exhibits notable distribution discrepancies among the various target domains, indicating its focus on extracting domain-specific knowledge.

\subsection{Distribution Distance}
To provide clearer evidence for our assumption, we directly calculate the distribution distance to represent different domain representation of adapters.
We adopt the domain distance definition proposed by Ben-David \citep{ben2006analysis, ben2010theory} and build upon previous domain transfer research \citep{ganin2016domain} by employing the $\mathcal{H}$-$divergence$ metric to further evaluate the domain representations of adapters across different target domains. $\mathcal{H}$-$divergence$ between $D_S$ and $D_{T_i}$ can be calculated as
\begin{equation}
d_\mathcal{H}(D_S, D_{T_i}) = 2 \mathop{\mathrm{sup}}_{\mathcal{D} \sim \mathcal{H}} | \mathop{\mathrm{Pr}}_{x \sim D_S}[\mathcal{D}(x)=1] - \mathop{\mathrm{Pr}}_{x \sim D_{T_i}}[\mathcal{D}(x)=1] |
\label{apeq:dh}
\end{equation}
, where $\mathcal{H}$ denotes hypothetical space and $\mathcal{D}$ denotes discriminator. 
Similar to \citep{ruder2017learning,allaway2021adversarial}, we adopt the $Jensen$-$Shannon\ (JS)\ divergence$ between two adjacent domains as an approximation of $\mathcal{H}$-divergence because it has been shown to successfully distinguish domains.
If the inter-domain divergence is relatively small, it can be demonstrated that the feature representation is consistent and less influenced by cross-domain shifts \citep{ganin2016domain}.
\begin{equation}
JS(P_{D_S} || P_{D_{T_i}}) = \frac{1}{2}KL(P_{D_S}|| \frac{P_{D_S}+P_{D_{T_i}}}{2}) + \frac{1}{2}KL(P_{D_{T_i}}|| \frac{P_{D_S}+P_{D_{T_i}}}{2})
\label{eq:js}
\end{equation}
Where $Kullback$-$Leibler\ (KL)\ divergence$ between two domain is

\begin{equation}
KL(P_1||P_2) = \sum_{i=0}^{n} P_1(x_i) log(\frac{P_1(x_i)}{P_2(x_i)})
\label{eq:kl}
\end{equation}

Where $P$ denotes probability distribution of model output features. We split the output feature space into mutually disjoint intervals $x_i$. $n$ range from 0 to 1000. To investigate the effectiveness of adapters in adapting to continual target domains, we compare the $JS$ values obtained by using the source model alone, injecting low-rank adapter, injecting high-rank adapter, and combining low-high adapters, as illustrated in Figure 3(a) of our submission. The low-rank adapter exhibits notably lower divergence values compared to the others, demonstrating robust task-relevant feature representation in various cross-domain phases. For high-rank adapter,  we use normalized intra-class divergence to further verify the domain representations of high-rank adapters in CIFAR10C, which is inspired by intra-cluster dissimilarity proposed by $k$-means~\citep{macqueen1967classification}. We first calculate the Euclidean distance clustering center for each category: 
\begin{equation}
\mu = \frac{1}{|C|} \sum_{e_i \sim C} e_i
\label{eq:kmeans}
\end{equation}
, where $e_i$ stands for output feature in class $C$. Then following \citep{macqueen1967classification}, we introduce normalized intra-class divergence $E$ by 
\begin{equation}
E = \phi(\frac{1}{|C|} \sum_{e_i \sim C} ||e_i - \mu||_2^2)
\label{apeq:kmeans}
\end{equation}
$\phi(\cdot)$ denotes for normlization function. In a given domain, if the intra-class divergence for each category is smaller, it demonstrates that the model has a better understanding of the current distribution \citep{li2020maximum}. As illustrated in Figure 3(b) of the submission, the high-rank adapter is found to drive down divergence within almost all domains and can better extract domain-specific knowledge in target domains.

\section{Additional Experiment}
\label{ap:ex}

\subsection{Additional Classification CTTA Experiments for Foundation Models}
\label{ap:clafo}

\begin{table*}[htb]
\caption{\label{tab:imageNet-to-imageNetc-dino} Average error rate (\%) for the ImageNet-to-ImageNet-C CTTA task. All results are evaluated on the ViT-Base, which uses the pre-trained encoder parameter of DINOv2 and SAM.}
\centering
\setlength\tabcolsep{2pt}
\begin{adjustbox}{width=1\linewidth,center=\linewidth}
\begin{tabular}{c|c|c|ccccccccccccccc|cc}
\toprule
Backbone&Method & REF &
 \rotatebox[origin=c]{70}{Gaussian} & \rotatebox[origin=c]{70}{shot} & \rotatebox[origin=c]{70}{impulse} & \rotatebox[origin=c]{70}{defocus} & \rotatebox[origin=c]{70}{glass} & \rotatebox[origin=c]{70}{motion} & \rotatebox[origin=c]{70}{zoom} & \rotatebox[origin=c]{70}{snow} & \rotatebox[origin=c]{70}{frost} & \rotatebox[origin=c]{70}{fog}  & \rotatebox[origin=c]{70}{brightness} & \rotatebox[origin=c]{70}{contrast} & \rotatebox[origin=c]{70}{elastic\_trans} & \rotatebox[origin=c]{70}{pixelate} & \rotatebox[origin=c]{70}{jpeg}
& Mean$\downarrow$ & Gain\\\midrule
\multirow{4}{*}{DINOv2}&Source & &52.3 & 50.5 & 51.2 & 57.3 & 83.8 & 60.1 & 62.6 &  47.1 & 56.9 & 58.1 & 22.5 & 88.4 & 60.3 & 32.4 & 35.0&54.6&0.0\\
&Tent \citep{DequanWangetal2021}&ICLR2021  &51.7 & 43.6 & 50.4 & 56.2 & 74.1 &  51.7 & 67.2 & 46.9 & 53.2 & 50.1 & 25.2 & 69.6 & 58.0 & 29.5 & 39.4& 51.1 &+3.5\\
&CoTTA \citep{wang2022continual}& CVPR2022 &51.4 & 62.1 & 50.4 & 78.3 & 75.2 & 62.8 & 60.3 & 48.4 & 59.0 & 58.8 & 31.6 & 90.7 & 49.2 & 39.1 & 36.5 & 56.9 &-2.3\\
&\cellcolor{green!10}\textbf{Ours} & Proposed &\cellcolor{green!10}\textbf{49.0}&\cellcolor{green!10}49.8&\cellcolor{green!10}50.7&\cellcolor{green!10}61.4&\cellcolor{green!10}\textbf{60.2}&\cellcolor{green!10}\textbf{49.7}&\cellcolor{green!10}\textbf{42.6}&\cellcolor{green!10}\textbf{47.1}&\cellcolor{green!10}\textbf{51.9}&\cellcolor{green!10}\textbf{45.3}&\cellcolor{green!10}27.1&\cellcolor{green!10}\textbf{49.7}&\cellcolor{green!10}\textbf{47.4}&\cellcolor{green!10}32.0&\cellcolor{green!10}\textbf{29.4}&\cellcolor{green!10}\textbf{46.2}&\cellcolor{green!10}+\textbf{8.4}\\ \hline
\multirow{4}{*}{SAM}&Source & &67.9 & 62.1 & 51.6 & 69.7 & 92.6 & 65.4 & 59.8 & 53.9 & 61.2 & 64.1 & 39.0 & 91.6 & 60.1 & 47.3 & 67.0 &63.6&0.0\\
&Tent \citep{DequanWangetal2021}&ICLR2021  &67.2 & 59.1 & 48.8 & 56.2 & 72.5 & 59.4 & 61.0 & 49.1 & 57.9 & 63.7 & 33.8 & 77.0 & 51.4 & 39.5 & 55.2 & 55.5 &+8.1\\
&CoTTA \citep{wang2022continual}& CVPR2022 &68.1 & 64.5 & 50.4 & 67.1 & 80.1 & 68.9 & 67.0 & 63.1 & 69.5 & 61.4 & 40.6 & 88.2 & 58.3 & 43.5 & 68.4 & 63.9& -0.3\\
&\cellcolor{green!10}\textbf{Ours} & Proposed &\cellcolor{green!10}\textbf{59.9}&\cellcolor{green!10}\textbf{55.7}&\cellcolor{green!10}\textbf{40.2}&\cellcolor{green!10}84.3&\cellcolor{green!10}\textbf{49.6}&\cellcolor{green!10}59.7&\cellcolor{green!10}\textbf{59.0}&\cellcolor{green!10}\textbf{47.8}&\cellcolor{green!10}\textbf{48.3}&\cellcolor{green!10}\textbf{57.4}&\cellcolor{green!10}\textbf{26.6}&\cellcolor{green!10}\textbf{71.8}&\cellcolor{green!10}\textbf{42.9}&\cellcolor{green!10}41.7&\cellcolor{green!10}\textbf{50.3}&\cellcolor{green!10}\textbf{53.0}&\cellcolor{green!10}+\textbf{10.6}\\ \hline
\bottomrule
\end{tabular}
\end{adjustbox}
\end{table*}

To demonstrate the effectiveness of our proposed method in enhancing the continual adaptation ability of foundation models such as DINOv2~\citep{oquab2023dinov2} and SAM~\citep{kirillov2023segment}, we conduct additional experiments on a more extensive dataset, namely ImagNet-to-ImageNet-C. Our approach involve loading the weight parameters of the foundation model and fine-tuning it on ImagNet, thus constructing our source model. It is important to note that we solely utilize the pre-trained encoder of SAM and incorporated a classification head, which is fine-tuned on the source domain. Subsequently, we adapt the source model to continual target domains (ImageNet-C) comprising fifteen corruption types.
The results, as depicted in Table~\ref{tab:imageNet-to-imageNetc-dino}, demonstrate that our approach achieved a significant performance improvement of 8.4\% on the representative image-level foundation model DINOv2 and 10.6\% on the pixel-level foundation model SAM. These outcomes underscore the effectiveness of our method for large-scale models, consistently and reliably improving performance across target domains. Combining Table 1-3 from the submission, we were surprised to discover a significant decrease in model performance for the classification CTTA task when using the pre-trained encoder parameters of SAM. As SAM is a pixel-level foundation model, we then attempted to investigate the effectiveness of SAM's pretrained parameters in the segmentation CTTA task.

\subsection{Additional Segmentation CTTA Experiments for Foundation Models}
\label{ap:segfo}
As shown in Table~\ref{aptab:samseg}, we conducted segmentation CTTA using SAM's pre-trained parameters on the Cityscapes-to-ACDC scenario. However, it's worth noting that the Segformer model \citep{xie2021segformer}, which we employed in our main experiments, does not incorporate positional encoding. Therefore, we adopted the SETR model \citep{zheng2021rethinking} as our new baseline for loading SAM's pre-trained parameters. As shown in the table, our approach with SAM's pre-trained parameters outperforms others on the ACDC target domains. This aligns with the assumption that SAM, being a pixel-level foundational model, excels in capturing fine-grained feature representations in dense CTTA tasks.

\begin{table*}[htb]
\centering
\setlength\tabcolsep{0.1cm}
\caption{\label{aptab:samseg} Performance comparison for Cityscape-to-ACDC CTTA. All results are evaluated on the SETR, which uses the pre-trained parameter of source model or SAM.}
\small
\begin{tabular}{c|c|cccc|c}
\toprule
Method & Pre-trained & Fog&	Night&	Rain&	Snow&	
Mean mIoU \\ \hline
Source \citep{xie2021segformer}& Source model&	72.6&	43.1&	63.0&	64.3&	60.8\\
Source \citep{xie2021segformer}& SAM \citep{kirillov2023segment}&	74.8&	44.1&	66.7&	66.6&	63.0\\
CoTTA \citep{wang2022continual}&SAM \citep{kirillov2023segment}&	75.4&	45.9&	67.3&	68.7&	64.3 \\
\textbf{Ours} &SAM \citep{kirillov2023segment}&	\textbf{76.5}&	\textbf{47.2}&	\textbf{68.1}&	\textbf{70.7}&	\textbf{65.6} \\
\bottomrule
\end{tabular}
\end{table*}

\subsection{Domain Generalization on a Different Number of Unseen Target Domains}
\label{ap:dg}

\begin{table*}[htb]
\centering
\setlength\tabcolsep{0.12cm}
 \caption{\label{tab:dg10} The domain generalization experiments on ImageNet-C, where the source model was continually adapted on the first 5 domains and directly tested on 10 unseen domains. The evaluation of the results was conducted using ViT-base.}
\small
\begin{tabular}{l|cccccccccc|c}
\toprule
 &  \multicolumn{10}{c|}{\textbf{Directly test on 10 unseen domains}}& \multicolumn{1}{c}{\textbf{Unseen}} \\ \hline
Method & \rotatebox[origin=c]{70}{motion} & \rotatebox[origin=c]{70}{zoom} & \rotatebox[origin=c]{70}{snow} & \rotatebox[origin=c]{70}{frost} & \rotatebox[origin=c]{70}{fog} & \rotatebox[origin=c]{70}{brightness} & \rotatebox[origin=c]{70}{contrast} & \rotatebox[origin=c]{70}{elastic\_trans} & \rotatebox[origin=c]{70}{pixelate} & \rotatebox[origin=c]{70}{jpeg}
& Mean$\downarrow$ \\\midrule
Source& 58.5&63.3&49.9&54.2&57.7&26.4&91.4&57.5&38.0&36.2&53.3\\
Tent \citep{DequanWangetal2021}& 56.0 & 61.3& 45.7 &49.6   &  56.6 &24.8&94.0&55.6&37.1&35.1&51.6\\
CoTTA \citep{wang2022continual}& 57.3 & 62.1& 49.1 &52.0   &  57.1 &26.4&91.9&57.1&37.6&35.3&52.6\\
\cellcolor{green!10}\textbf{Ours} & \cellcolor{green!10}\textbf{46.4} &\cellcolor{green!10}\textbf{52.7} &\cellcolor{green!10}\textbf{39.8}  & \cellcolor{green!10}\textbf{43.7}  &\cellcolor{green!10}\textbf{42.2}   &\cellcolor{green!10}\textbf{23.5}&\cellcolor{green!10}\textbf{71.5}&\cellcolor{green!10}\textbf{49.6}&\cellcolor{green!10}\textbf{33.9}&\cellcolor{green!10}\textbf{33.3}&\cellcolor{green!10}\textbf{43.7}\\ 
\bottomrule
\end{tabular}
\end{table*}

\begin{table*}[htb]
\centering
\setlength\tabcolsep{0.21cm}
 \caption{\label{tab:dg8} The domain generalization experiments on ImageNet-C, where the source model was continually adapted on the first 7 domains and directly tested on 8 unseen domains. The evaluation of the results was conducted using ViT-base.}
\small
\begin{tabular}{l|cccccccc|c}
\toprule
 &  \multicolumn{8}{c|}{\textbf{Directly test on 8 unseen domains}}& \multicolumn{1}{c}{\textbf{Unseen}} \\ \hline
Method  & \rotatebox[origin=c]{70}{snow} & \rotatebox[origin=c]{70}{frost} & \rotatebox[origin=c]{70}{fog} & \rotatebox[origin=c]{70}{brightness} & \rotatebox[origin=c]{70}{contrast} & \rotatebox[origin=c]{70}{elastic\_trans} & \rotatebox[origin=c]{70}{pixelate} & \rotatebox[origin=c]{70}{jpeg}
& Mean$\downarrow$ \\\midrule
Source& 49.9  &54.2   &57.7   &26.4&91.4&57.5&38.0&36.2&51.4\\
Tent \citep{DequanWangetal2021} &44.3  &48.8   &51.8   &24.9&83.7&55.2&35.4&34.7&47.4\\
CoTTA \citep{wang2022continual} & 48.8 & 52.2  &56.7   &26.1&91.1&57.0&37.3&35.3&50.6\\
\cellcolor{green!10}\textbf{Ours} &\cellcolor{green!10}\textbf{39.6}  & \cellcolor{green!10}\textbf{43.7}  &\cellcolor{green!10}\textbf{41.7}   &\cellcolor{green!10}\textbf{23.7}&\cellcolor{green!10}\textbf{63.7}&\cellcolor{green!10}\textbf{51.7}&\cellcolor{green!10}\textbf{33.3}&\cellcolor{green!10}\textbf{33.6}&\cellcolor{green!10}\textbf{42.3}\\ 
\bottomrule
\end{tabular}
\end{table*}

Similar to our previous submission, we follow the leave-one-domain-out principle \citep{zhou2021domain, li2017deeper}, where we utilize a subset of ImageNet-C domains as new source domains for model training, while leaving the remaining domains as target domains without any adaptation. However, in contrast to previous domain generalization experiments, we adopt an unsupervised continual test-time adaptation (CTTA) approach for training the model on these unlabeled source domains. We solely utilize the ImageNet pre-trained parameters as the initial weights of the model.
In the supplementary material, we utilize 5 out of 15 and 7 out of 15 domains from ImageNet-C as the source domains, leaving the remaining 10 out of 15 and 8 out of 15 domains as unseen target domains.
Surprisingly, the results presented in Table \ref{tab:dg10} and \ref{tab:dg8} demonstrate that our method achieves a reduction of 9.6\% and 9.1\% in the average error on these unseen domains, respectively. These promising outcomes validate the DG ability of our method, as it effectively extracts domain-shared knowledge and provides a new perspective for enhancing DG performance within an unsupervised paradigm.

\subsection{Additional Ablation study}
\label{ap:ab}

\textbf{How does the middle-layer dimension influence the performance?}

\begin{figure*}[htb]
\centering
\includegraphics[width=0.65\linewidth]{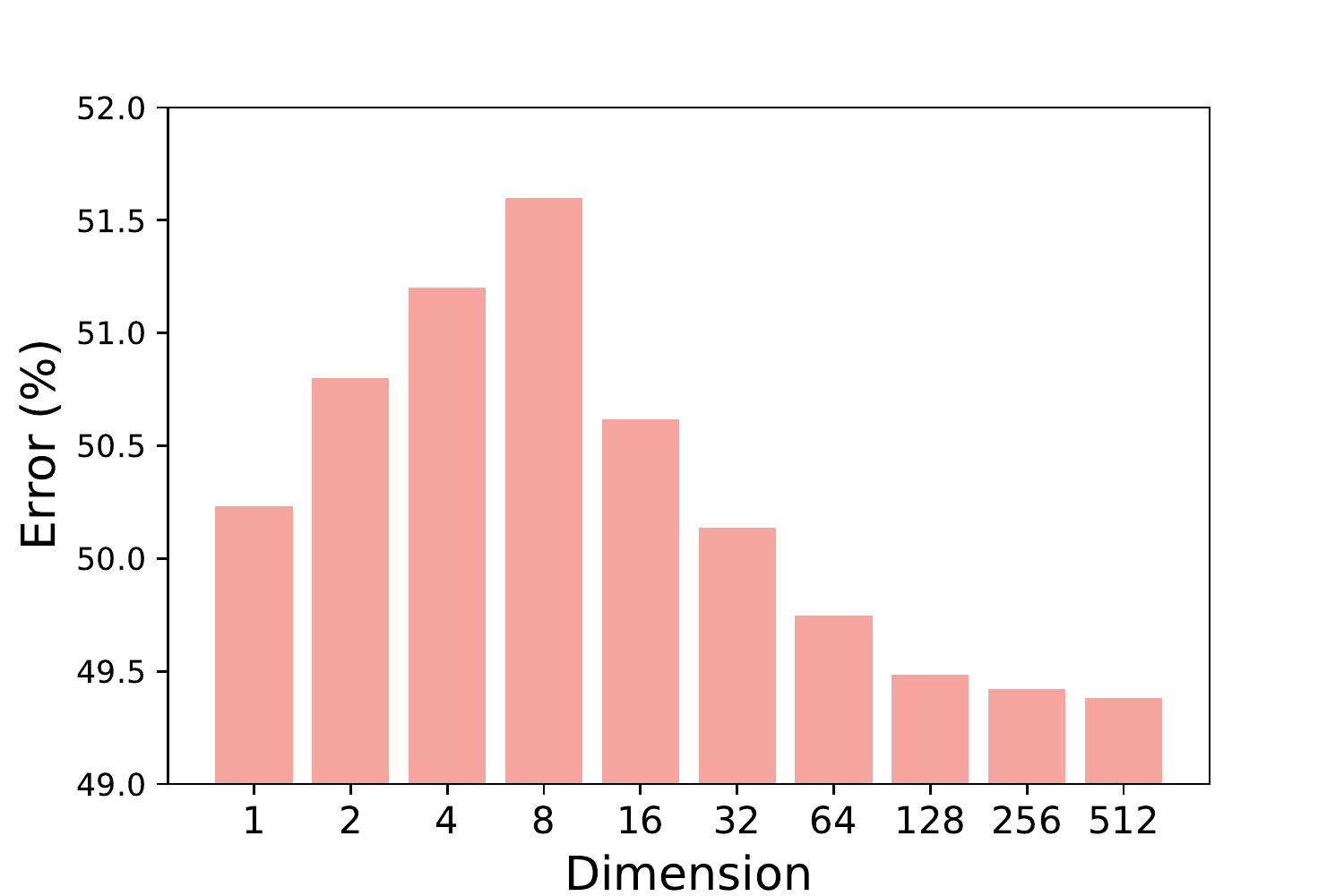}
\vspace{-0.15cm}
\caption{The middle-layer dimension influence of the performance}
\label{fig:rank}
\end{figure*}

According to Figure \ref{fig:rank}, we observe that as the dimension decreases, the error rate concurrently drops. This trend suggests that lower-dimension middle layer more effectively extract the domain-shared knowledge, leading to an improved model performance. However, an opposite trend emerges when dimension surpasses 16, with performance enhancements accompanying increased dimension. This correlation implies that middle layers with a higher dimension excel in extracting domain-specific knowledge. And we find that when the dimension is larger than 128, the performance improvement is limited but brings a larger number of parameters. Therefore, we set the dimension of the high-dimension middle layer to 128 in our study.

\textbf{How do different adapter initialization methods impact ViDA performance?}

Pre-training the low-rank and high-rank ViDAs using source data is an unnecessary step and does not compromise the effectiveness of our approach. ViDAs can demonstrate comparable CTTA performance when they have a relatively stable initial parameter. As illustrated in the Table \ref{aptab:initial}, we conduct an additional experiment on the Cityscape-to-ACDC scenario. ViDAs with random initial parameters and ViDAs with parameters pre-trained on ImageNet achieved 60.5 and 61.4 mIoU in target domains, respectively, exhibiting notable improvements compared to previous methods.

\begin{table*}[htb]
\caption{\label{aptab:initial}The ablation study examines adapter initialization methods on the Cityscape-to-ACDC CTTA scenario.}
\small
\centering
\setlength\tabcolsep{6pt}
\begin{tabular}{c|c|cccc|c}
\toprule
  & Adapter pre-train & Fog & Night & Rain & Snow & Mean (IoU) \\\midrule
 Source & - & 69.1 & 40.3 & 59.7 & 57.8 & 56.7 \\
CoTTA & - & 70.9 & 41.2 & 62.4 & 59.7 & 58.6 \\
Ours & Source & 71.6 & 43.2 & 66.0 & 63.4 & 61.1\\
Ours & Random initial & 71.6& 43.6 & 64.9 & 61.9 & 60.5 \\
Ours & ImageNet &71.6&44.3&66.0&63.5&61.4\\ \hline
\end{tabular}
\end{table*}

\subsection{Experiments on Classification CTTA with Convolutional Backbones}
\label{ap:conv}

\begin{table*}[htb]
\caption{\label{aptab:wide}Classification error rate(\%) for standard CIFAR10-to-CIAFAR10C online CTTA task. Results are evaluated on WideResNet-28. Mean is the average value of the error rate. Gain(\%) represents the percentage of improvement in model accuracy compared with the source method.}
\small
\centering
\setlength\tabcolsep{2pt}
\begin{tabular}{c|c|c|cc}
\toprule
Method & REF & Conference
& Mean$\downarrow$ & Gain\\\midrule
Source &\citep{SergeyZagoruykoetal2016}&BMVC2016&43.5&0.0\\
BN Stats Adapt &\citep{Schneideretal2020}&NeurIPS2020&20.4&+23.1\\
TENT  &\citep{DequanWangetal2021}&ICLR2021&20.7&+22.8\\
CoTTA &\citep{wang2022continual}&CVPR2022&16.2&+27.3\\
RoTTA& \citep{yuan2023robust}&CVPR2023& 17.5 & +26.0 \\
NOTE&\citep{gong2022note}&NeurIPS2022& 20.2 & +23.3\\
EcoTTA& \citep{song2023ecotta}&ICCV2023& 16.8 & +26.7\\
SATA & \citep{chakrabarty2023sata}&2023.4.20 &16.1&+27.4\\
\cellcolor{green!10}\textbf{Ours } & Proposed&2023.5.18&\cellcolor{green!10}\textbf{15.8}&\cellcolor{green!10}+\textbf{27.7}\\ \hline
\bottomrule
\end{tabular}
\end{table*}

\noindent\textbf{CIFAR10-to-CIFAR10C standard task.}
In contrast to the experiments conducted in our submission, we introduce a change in the backbone of the classification model to WideResNet-28, which is consistent with previous works \citep{wang2022continual}. Specifically, we modify the up-projection layer and down-projection layer to utilize $1 \times 1$ convolutions, while the adapters are placed alongside the original $3 \times 3$ convolutions. For ViDA, we maintain a low-rank dimension of 1 and a high-rank dimension of 128.
As depicted in Table \ref{aptab:wide}, our method achieves a 27.7\% improvement over the source model. These findings demonstrate that our method successfully address error accumulation and catastrophic forgetting problem, regardless of the network backbone employed.

\subsection{Additional Experiments on Segmentation CTTA}
\label{ap:seg}

\begin{table*}[htb]
\caption{\label{aptab:ACDC} \textbf{10 rounds segmentation CTTA on Cityscape-to-ACDC.} We sequentially repeat the same sequence of target domains 10 times. Mean is the average score of mIoU.}
\centering
\setlength\tabcolsep{0.5pt}
\begin{adjustbox}{width=1\linewidth,center=\linewidth}
\begin{tabular}{c|ccccc|ccccc|ccccc|ccccc|ccccc|c }
\hline

Round       & \multicolumn{5}{c|}{1}    & \multicolumn{5}{c|}{2}     & \multicolumn{5}{c|}{3} & \multicolumn{5}{c|}{4} & \multicolumn{5}{c|}{5}  & Mean   \\ \cline{1-27}
Method  & Fog & Night & Rain & Snow & Mean & Fog & Night & Rain & Snow  & Mean & Fog & Night & Rain & Snow & Mean &   Fog & Night & Rain & Snow & Mean &  Fog & Night & Rain & Snow & Mean \\ \hline
Source   &69.1&40.3&59.7&57.8&56.7&69.1&40.3&59.7& 	57.8&56.7&69.1&40.3&59.7& 57.8&56.7&56.7 &40.3&59.7& 57.8&56.7&56.7 &40.3&59.7& 57.8&56.7&cont. \\ 
CoTTA  &70.9&41.2&62.4&59.7&58.6&70.9&41.1&62.6& 	59.7&58.6&70.9&41.0&62.7&59.7&58.6&70.9&41.0&62.7&59.7&58.6&70.9&41.0&62.8&59.7&58.6&cont.\\ 
CoTTA$^{*}$  &71.9&45.0& 67.1&63.1&61.8&71.9&43.6&65.6& 	61.8&60.7&69.6&39.7&63.5&60.4&58.3&68.3&39.6&61.8&59.4&57.3&67.8&38.9&62.1&59.7&57.1&cont.\\
Ours  & 71.6& 43.2& 66.0& 63.4& \textbf{61.1}& 
  73.2& 44.5& 67.0& 63.9& \textbf{62.2} 
 & 73.2& 44.6& 67.2& 64.2& \textbf{62.3}  &70.9&44.0&66.0&63.2&\textbf{61.0}&72.0&43.7&66.3&63.1&\textbf{61.3}&cont.\\ \hline

Round       & \multicolumn{5}{c|}{6}    & \multicolumn{5}{c|}{7}     & \multicolumn{5}{c|}{8} & \multicolumn{5}{c|}{9} & \multicolumn{5}{c|}{10}  & Mean   \\ \cline{1-27}
Method  & Fog & Night & Rain & Snow & Mean & Fog & Night & Rain & Snow  & Mean & Fog & Night & Rain & Snow & Mean &   Fog & Night & Rain & Snow & Mean &  Fog & Night & Rain & Snow & Mean \\ \hline
Source    &69.1&40.3&59.7&57.8&56.7&69.1&40.3&59.7& 	57.8&56.7&69.1&40.3&59.7& 57.8&56.7&56.7 &40.3&59.7& 57.8&56.7&56.7 &40.3&59.7& 57.8&56.7&56.7 \\ 
CoTTA  &70.9&41.0&62.8&59.7&58.6&70.9&41.1&62.6& 	59.7&58.6&70.9&41.1&62.6& 	59.7&58.6&70.8&41.1&62.6& 	59.7&58.6&70.8&41.1&62.6& 	59.7&58.6&58.6 \\ 
CoTTA$^{*}$  &67.7&39.8& 62.7&59.7&57.5&67.3&39.7&63.2&59.6&57.7&67.6&40.1&63.2&58.0&57.2&65.0&38.8&60.7&58.5&55.8&66.9&38.9&62.7&58.7&56.8&58.0\\
Ours   &  72.2&44.0&66.6&62.9&\textbf{61.4}&72.3&44.8&66.5&62.9& 	\textbf{61.6}&72.1&45.1&66.2&62.9&\textbf{61.5}&71.9&45.3&66.3&62.9&\textbf{61.5}&72.2&45.2&66.5&62.9&\textbf{61.6}&\textbf{61.6}\\
\bottomrule
 \hline
\end{tabular}
\end{adjustbox}
\end{table*}

We further present the segmentation CTTA experiment with 10 rounds on Table \ref{aptab:ACDC}. Notably, it demonstrates a consistent enhancement in mean mIoU during the initial rounds (rounds 1-3) while maintaining stable performance in subsequent rounds (rounds 4-10). After averaging over 10 rounds , our method achieved a 3.0\% mIoU improvement compared to the previous SOTA method. 
As shown in Table \ref{aptab:ACDC} (CoTTA$^{*}$), we adjust the hyperparameters of the CoTTA method by raising the learning rate to 3e-4, which aligns with our implementation details. The impact of this adjustment is evident in the initial three rounds of segmentation, where performance notably improves. However, as we progress to subsequent CTTA rounds, we observe a noticeable decline in segmentation accuracy and encounter the problem of catastrophic forgetting.

\section{Additional Qualitative Analysis}
\label{ap:qu}

\begin{figure*}[htb]
\centering
\includegraphics[width=\linewidth]{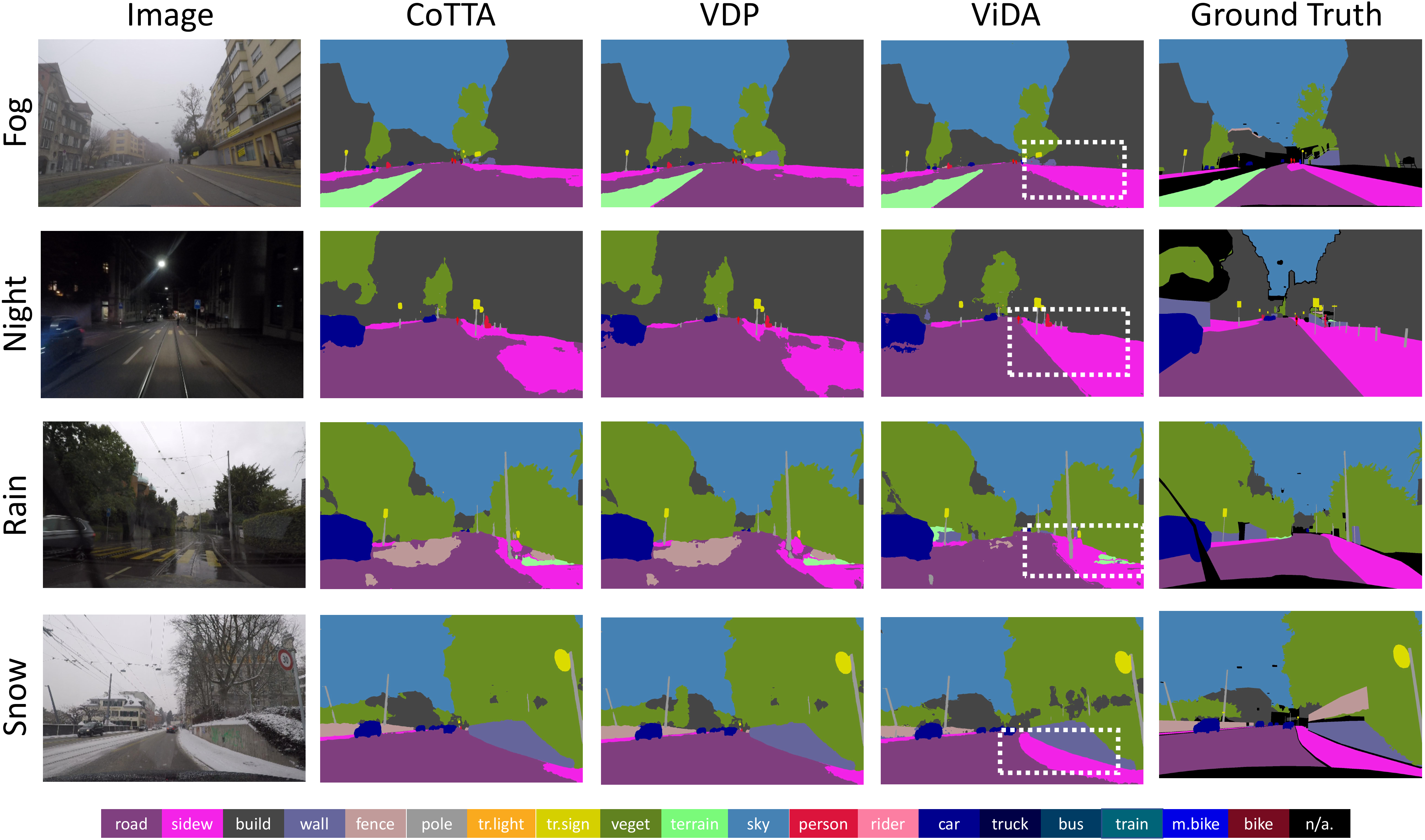}
\caption{Qualitative comparison of our method with previous SOTA methods on the ACDC dataset. Our method could better segment different pixel-wise classes such as shown in the white box.}
\label{fig:vis_sup}
\end{figure*}

To further validate the effectiveness of our proposed method, we present additional qualitative comparisons on the Cityscapes-to-ACDC CTTA scenario. Initially, we pre-train the Segformer-B5 model \citep{xie2021segformer} on the source domain and subsequently adapt it to four target domains in ACDC. In order to assess the performance of our approach, we conduct a qualitative comparison with two leading methods, namely CoTTA \citep{wang2022continual} and VDP \citep{gan2023decorate}. The visualizations of the segmentation outputs, obtained through the CTTA process, are depicted in Figure \ref{fig:vis_sup}.
Our method exhibits better segmentation map compared to CoTTA and VDP across all four target domains, as it effectively distinguishes the sidewalk from the road (shown in white box). This demonstrates the capability of our method to achieve more accurate segmentation results while mitigating the impact of dynamic domain shifts. Moreover, in the other categories, our method's segmentation maps closely resemble the Ground Truth, leading to a visual enhancements. Lastly, we have included a video visualization in the supplementary material that showcases a comprehensive comparison of segmentation performance. This video provides a dynamic and visual representation of the results obtained from our experiments.

\section{Fine-grained Performance}
\label{ap:fi}

\begin{table*}[htb]
\caption{\label{tab:CIFAR10-to-CIFAR10C-sup} A fine-grained Classification error rate(\%) for standard CIFAR10-to-CIAFAR10C online CTTA task. Results are evaluated on ViT-base.}
\centering
\setlength\tabcolsep{2pt}
\begin{adjustbox}{width=1\linewidth,center=\linewidth}
\begin{tabular}{l|ccccccccccccccc|cc}
\toprule
Method & 
 \rotatebox[origin=c]{70}{gaussion} & \rotatebox[origin=c]{70}{shot} & \rotatebox[origin=c]{70}{impulse} & \rotatebox[origin=c]{70}{defocus} & \rotatebox[origin=c]{70}{glass} & \rotatebox[origin=c]{70}{motion} & \rotatebox[origin=c]{70}{zoom} & \rotatebox[origin=c]{70}{snow} & \rotatebox[origin=c]{70}{frost} & \rotatebox[origin=c]{70}{fog} & \rotatebox[origin=c]{70}{bri.} & \rotatebox[origin=c]{70}{contrast} & \rotatebox[origin=c]{70}{elastic\_trans} & \rotatebox[origin=c]{70}{pixelate} & \rotatebox[origin=c]{70}{jpeg}
& Mean$\downarrow$ & Gain\\\midrule
Source&60.1&53.2&38.3&19.9&35.5&22.6&18.6&12.1&12.7&22.8&5.3&49.7&23.6&24.7&23.1&28.2&0.0\\
Pseudo-label \citep{Leeetal2013}&59.8&52.5&37.2&19.8&35.2&21.8&17.6&11.6&12.3&20.7&5.0&41.7&21.5&25.2&22.1&26.9&+1.3\\
TENT-continual \citep{DequanWangetal2021}&57.7&56.3&29.4&16.2&35.3&16.2&12.4&11.0&11.6&14.9&4.7&22.5&15.9&29.1&19.5&23.5&+4.7\\
CoTTA \citep{wang2022continual}&58.7&51.3&33.0&20.1&34.8&20&15.2&11.1&11.3&18.5&4.0&34.7&18.8&19.0&17.9&24.6&+3.6\\
VDP\citep{gan2023decorate}&57.5&49.5&31.7&21.3&35.1&19.6&15.1&10.8&10.3&18.1&4&27.5&18.4&22.5&19.9&24.1&+4.1\\
\cellcolor{green!10}\textbf{Ours (proposed)} &\cellcolor{green!10}\textbf{52.9}&\cellcolor{green!10}\textbf{47.9}&\cellcolor{green!10}\textbf{19.4}&\cellcolor{green!10}\textbf{11.4}&\cellcolor{green!10}\textbf{31.3}&\cellcolor{green!10}\textbf{13.3}&\cellcolor{green!10}\textbf{7.6}&\cellcolor{green!10}\textbf{7.6}&\cellcolor{green!10}\textbf{9.9}&\cellcolor{green!10}\textbf{12.5}&\cellcolor{green!10}\textbf{3.8}&\cellcolor{green!10}\textbf{26.3}&\cellcolor{green!10}\textbf{14.4}&\cellcolor{green!10}\textbf{33.9}&\cellcolor{green!10}\textbf{18.2}&\cellcolor{green!10}\textbf{20.7}&\cellcolor{green!10}+\textbf{7.5}\\
\bottomrule
\end{tabular}
\end{adjustbox}
\end{table*}

\begin{table*}[htb]
\caption{\label{tab:CIFAR100-to-CIFAR100C-sup}A fine-grained Classification error rate(\%) for standard CIFAR100-to-CIAFAR100C online CTTA task. Results are evaluated on ViT-base.}
\centering
\setlength\tabcolsep{2pt}
\begin{adjustbox}{width=1\linewidth,center=\linewidth}
\begin{tabular}{l|ccccccccccccccc|cc}
\toprule
Method & \rotatebox[origin=c]{70}{gaussion} & \rotatebox[origin=c]{70}{shot} & \rotatebox[origin=c]{70}{impulse} & \rotatebox[origin=c]{70}{defocus} & \rotatebox[origin=c]{70}{glass} & \rotatebox[origin=c]{70}{motion} & \rotatebox[origin=c]{70}{zoom} & \rotatebox[origin=c]{70}{snow} & \rotatebox[origin=c]{70}{frost} & \rotatebox[origin=c]{70}{fog} & \rotatebox[origin=c]{70}{bri.} & \rotatebox[origin=c]{70}{contrast} & \rotatebox[origin=c]{70}{elastic\_trans} & \rotatebox[origin=c]{70}{pixelate} & \rotatebox[origin=c]{70}{jpeg}
& Mean$\downarrow$ & Gain\\\midrule
Source&55.0&51.5&26.9&24.0&60.5&29.0&21.4&21.1&25.0&35.2&11.8&34.8&43.2&56.0&35.9&35.4&0.0\\
Pseudo-label \citep{Leeetal2013}&53.8&48.9&25.4&23.0&58.7&27.3&19.6&20.6&23.4&31.3&11.8&28.4&39.6&52.3&33.9&33.2&+2.2\\
TENT-continual  \citep{DequanWangetal2021} &53.0&47.0&24.6&22.3&58.5&26.5&19.0&21.0&23.0&30.1&11.8&25.2&39.0&47.1&33.3&32.1&+3.3\\
CoTTA  \citep{wang2022continual}&55.0&51.3&25.8&24.1&59.2&28.9&21.4&21.0&24.7&34.9&11.7&31.7&40.4&55.7&35.6&34.8&+0.6\\
VDP \citep{gan2023decorate}&54.8&51.2&25.6&24.2&59.1&28.8&21.2&20.5&23.3&33.8&7.5&11.7&32.0&51.7&35.2&32.0&+3.4\\
\cellcolor{green!10}\textbf{Ours (proposed)} &\cellcolor{green!10}\textbf{50.1}&\cellcolor{green!10}\textbf{40.7}&\cellcolor{green!10}\textbf{22.0}&\cellcolor{green!10}\textbf{21.2}&\cellcolor{green!10}\textbf{45.2}&\cellcolor{green!10}\textbf{21.6}&\cellcolor{green!10}\textbf{16.5}&\cellcolor{green!10}\textbf{17.9}&\cellcolor{green!10}\textbf{16.6}&\cellcolor{green!10}\textbf{25.6}&\cellcolor{green!10}\textbf{11.5}&\cellcolor{green!10}\textbf{29.0}&\cellcolor{green!10}\textbf{29.6}&\cellcolor{green!10}\textbf{34.7}&\cellcolor{green!10}\textbf{27.1}&\cellcolor{green!10}\textbf{27.3}&\cellcolor{green!10}+\textbf{8.1}\\ 
\bottomrule
\end{tabular}
\end{adjustbox}
\vspace{-0.2cm}
\end{table*}

In this section, we expand upon the classification results presented in our submission by providing a details of fine-grained performance. We assess the error rates across fifteen corruption types to gain deeper insights. To be specific, we augment the information provided in Table 2 of our submission with the additional details presented in Table \ref{tab:CIFAR10-to-CIFAR10C-sup} and \ref{tab:CIFAR100-to-CIFAR100C-sup}. These tables offer a comprehensive view of the performance of our approach in addressing the CIFAR-10-to-CIFAR-10C and CIFAR-100-to-CIFAR-100C CTTA scenarios, respectively.

\end{document}